\documentclass[11pt]{article}

\usepackage[a4paper,margin=1in]{geometry}
\usepackage{amsmath,amssymb,amsthm,mathtools}
\usepackage{microtype}
\usepackage{booktabs}
\usepackage[dvipsnames]{xcolor}
\usepackage[authoryear,round]{natbib}
\usepackage[hidelinks,pagebackref=true]{hyperref}
\expandafter\ifx\csname iflatexml\endcsname\relax\expandafter\newif\csname iflatexml\endcsname\fi
\iflatexml\newcommand{\faGithub}{}\else\usepackage{fontawesome5}\fi

\usepackage{hyperref}
\usepackage[dvipsnames]{xcolor}
\definecolor{BlueIMT}{RGB}{0,42,72}
 \hypersetup{
    colorlinks=true,
    urlcolor=BlueIMT,
    linkcolor=MidnightBlue,
    citecolor=MidnightBlue
 }

\usepackage{enumitem}
\usepackage{float}

\usepackage[capitalize,noabbrev]{cleveref}

\usepackage{crossreftools}
\usepackage{xargs} 
\usepackage{xstring}
\usepackage{etoolbox}

\let\defaulturl\url
\newcommand{\biburl}[1]{\href{#1}{[url]}}
\AtBeginEnvironment{thebibliography}{%
  \let\url\biburl
}
\AtEndEnvironment{thebibliography}{%
  \let\url\defaulturl
}

\renewcommand*{\backrefalt}[4]{%
  \ifcase #1\relax
  \or
    (cited on p.~#2)%
  \else
    (cited on pp.~#2)%
  \fi
}

\usepackage{etoc}
\usepackage{titletoc}
\usepackage[bibliography=common]{apxproof}
\usepackage{aliascnt}

\newtheoremrep{thm}{Theorem}[section]
\newaliascnt{cor}{thm}

\aliascntresetthe{cor}
\newaliascnt{lemma}{thm}
\newtheorem{lemma}[lemma]{Lemma}
\aliascntresetthe{lemma}
\newaliascnt{fact}{thm}

\aliascntresetthe{fact}
\newaliascnt{prop}{thm}

\aliascntresetthe{prop}

\newaliascnt{definition}{thm}
\newtheorem{definition}[definition]{Definition}
\aliascntresetthe{definition}

\theoremstyle{remark}
\newaliascnt{remark}{thm}
\newtheorem{remark}[remark]{Remark}
\aliascntresetthe{remark}

\crefname{thm}{theorem}{theorems}
\crefname{assumption}{assumption}{assumptions}
\Crefname{assumption}{Assumption}{Assumptions}
\crefname{cor}{corollary}{corollaries}
\Crefname{cor}{Corollary}{Corollaries}
\crefname{prop}{proposition}{propositions}
\Crefname{prop}{Proposition}{Propositions}
\crefname{lemma}{lemma}{lemmas}
\Crefname{lemma}{Lemma}{Lemmas}
\crefname{fact}{fact}{facts}
\crefname{definition}{definition}{definitions}
\Crefname{definition}{Definition}{Definitions}
\crefname{remark}{remark}{remarks}
\Crefname{remark}{Remark}{Remarks}

\usepackage{mdframed}
\newmdtheoremenv{algo}{Algorithm}
\newmdtheoremenv{procedure}{Decision process}

\newlist{assnum}{enumerate}{1} %
\setlist[assnum]{label=(\roman*), ref=\theassumption(\roman*)}
\crefalias{assnumi}{assumption} 

\newlist{lemnum}{enumerate}{1} %
\setlist[lemnum]{label=(\roman*), ref=\thelemma(\roman*)}
\crefalias{lemnumi}{lemma} 

\newlist{thmnum}{enumerate}{1} %
\setlist[thmnum]{label=(\roman*), ref=\thethm(\roman*)}
\crefalias{thmnumi}{thm} 

\newlist{cornum}{enumerate}{1} %
\setlist[cornum]{label=(\roman*), ref=\thecor(\roman*)}
\crefalias{cornumi}{cor} 

\newlist{definitionnum}{enumerate}{1} %
\setlist[definitionnum]{label=(\roman*), ref=\thedefinition(\roman*)}
\crefalias{definitionnumi}{definition} 

\newlist{propnum}{enumerate}{1} %
\setlist[propnum]{label=(\roman*), ref=\theproposition(\roman*)}
\crefalias{propnumi}{prop}

\newlist{examplenum}{enumerate}{1} %
\setlist[examplenum]{label=(\roman*), ref=\theexample(\roman*)}
\crefalias{examplenumi}{example}

\usepackage{dsfont}
\usepackage{nicefrac}

\newcommand{\lmo}{\mathsf{lmo}}

\usepackage{tcolorbox}
\newtcolorbox{defbox}{colback=black!5!white,colframe=black!75!black}
\newtcolorbox{asmbox}{colback=black!5!white,colframe=black!75!black}
\newtcolorbox{thmbox}{colback=red!5!white,colframe=red!75!black}

\usepackage{braket}

\newcommandx{\QC}[2][1={},2={}]{\ifstrempty{#1}{Q#2}{Q_{#1}\ifstrempty{#2}{}{(#2)}}}
\newcommandx{\PC}[2][1={},2={}]{\ifstrempty{#1}{P#2}{P_{#1}\ifstrempty{#2}{}{(#2)}}}
\newcommandx{\HC}[2][1={},2={}]{\ifstrempty{#1}{H#2}{H_{#1}\ifstrempty{#2}{}{(#2)}}}
\newcommandx{\MC}[2][1={},2={}]{\ifstrempty{#1}{M#2}{M_{#1}\ifstrempty{#2}{}{(#2)}}}

\newcommandx{\EF}[2][1={k},2={}]{\mathbb E\ifstrempty{#1}{}{_{#1}}#2}

\newtcolorbox[auto counter, number within=section,crefname={insight}{insights}]{insightbox}[1][]{%
    colback=MidnightBlue!5!white,
    colframe=MidnightBlue!40!white,
    sharp corners,
    before upper={{\bfseries\color{MidnightBlue}Insight~\thetcbcounter}.\ },
    #1
}

\newtcolorbox[auto counter, number within=section,crefname={insight}{insights}]{recommendationbox}[1][]{%
    colback=green!5!white,
    colframe=green!15!white,
    sharp corners,
    before upper={{\bfseries Recommendation~\thetcbcounter}.\ },
    #1
}

\usepackage{algorithm}
\usepackage[noend]{algpseudocode}

\newcommand{\equationbox}[1]{\par\noindent\fbox{%
\begin{minipage}{\dimexpr\linewidth-2\fboxsep-2\fboxrule\relax}
#1
\end{minipage}}\par}

\usepackage{adjustbox}

\usepackage{array,multirow,hhline,graphicx}
\usepackage{wrapfig}
\usepackage{colortbl}

\usepackage{pifont}

\usepackage[para,online,flushleft]{threeparttable}

\theoremstyle{plain}
\newtheorem{theorem}[thm]{Theorem}
\newaliascnt{proposition}{thm}
\newtheorem{proposition}[proposition]{Proposition}
\aliascntresetthe{proposition}
\crefname{proposition}{proposition}{propositions}
\Crefname{proposition}{Proposition}{Propositions}
\newaliascnt{corollary}{thm}

\aliascntresetthe{corollary}
\crefname{corollary}{corollary}{corollaries}
\Crefname{corollary}{Corollary}{Corollaries}
\theoremstyle{definition}
\newtheorem{desideratum}{Desideratum}
\crefname{desideratum}{desideratum}{desiderata}
\Crefname{desideratum}{Desideratum}{Desiderata}
\theoremstyle{remark}

\newcommand{\Fix}{\operatorname{Fix}}
\newcommand{\cC}{\mathcal{C}}
\newcommand{\opA}{\mathbb{A}}
\newcommand{\opB}{\mathbb{B}}
\newcommand{\opC}{\mathbb{C}}
\newcommand{\opD}{\mathbb{D}}
\newcommand{\opE}{\mathbb{E}}
\newcommand{\opF}{\mathbb{F}}
\newcommand{\opG}{\mathbb{G}}
\newcommand{\opH}{\mathbb{H}}
\newcommand{\opI}{\mathbb{I}}
\newcommand{\opJ}{\mathbb{J}}
\newcommand{\opK}{\mathbb{K}}
\newcommand{\opM}{\mathbb{M}}
\newcommand{\opN}{\mathbb{N}}
\newcommand{\opP}{\mathbb{P}}
\newcommand{\opQ}{\mathbb{Q}}

\newcommand{\opS}{\mathbb{S}}
\newcommand{\opT}{\mathbb{T}}
\newcommand{\opU}{\mathbb{U}}
\newcommand{\opV}{\mathbb{V}}
\newcommand{\opW}{\mathbb{W}}
\newcommand{\opZ}{\mathbb{Z}}

\newcommand{\Sparrow}{\hyperref[alg:sparrow]{\textsf{Sparrow}}}

\newcommand{\budgetmult}{\zeta}

\title{How to build and train your recurrent-depth model: \\ A fixed-point perspective}
\author{Thomas Pethick\\\href{mailto:tmpethick@gmail.com}{\textcolor{MidnightBlue}{\texttt{tmpethick@gmail.com}}} \vspace{-0.5em}}
\date{}

\begin{document}
\maketitle

\begin{abstract}
Recurrent-depth models are trained with truncated backpropagation, yet it remains unclear why this works.
We show that input injection, commonly used in recurrent-depth models, is exactly Halpern iteration, and that a slowly vanishing stepsize provably converges with a backpropagation window polylogarithmic in the depth.
The guarantees require a Lipschitz operator, which for SwiGLU leads to a weight constraint in both row and spectral norm, while only a row norm constraint for gates.
Based on these constraints, we develop an optimizer tailored to each parameter group, $\textsf{Sparrow}$,  which provides a drop-in replacement of Muon/MuonH with no overhead, that speeds up training even beyond recurrent models.
\end{abstract}

{\centering\href{https://github.com/tmpethick/sparrow}{\textcolor{magenta}{\faGithub\ \texttt{github.com/tmpethick/sparrow}}}\par}

\etocdepthtag.toc{mainmatter}
\section{Introduction}
Recurrent-depth models such as the latent reasoning model Huginn-0125~\citep{geiping2025} scale test-time computation by repeatedly applying a shared block to a latent state.
Remarkably, they are trained with \emph{truncated} backpropagation, differentiating through only the last few iterations of the recurrence.

At first sight it seems almost magical that this works.
The truncated gradient ignores how all earlier iterations depend on the parameters, and nothing a priori prevents the latent state from drifting as the number of iterations grows.
One component that appears to be important in these architectures is \emph{input injection}, which feeds the encoded query back into every iteration~\citep{bansal2022,geiping2025}.

In a seemingly separate vein, constraining or normalizing the \emph{rows} of weight matrices has recently become widespread in both architectures and optimizers such as nGPT, NorMuon, Aurora and Muown~\citep{loshchilov2025ngpt,li2025normuon,dewulf2026aurora,lion2026muown}, often in combination with the spectral updates of Muon~\citep{jordan2024muon}.
It is less clear when such row constraints help, and on which matrices they should be imposed.

As we will see, these two threads are closely connected, a connection that will be revealed by asking the following question:
\begin{center}
\itshape
How should we structure the forward pass for truncated backpropagation to provably converge?\\
How should we train the resulting model?
\end{center}

To reason about these questions, we treat the recurrence as a fixed-point problem.
The model takes in some encoded query $u$ and iterates a learnable refinement map, the forward operator $\opT_\theta$, multiple times.
We can extract an unreasonable amount from this perspective.

First, it turns out that additive input injection is exactly an instance of Halpern iteration~\citep{halpern1967} with a fixed stepsize.
The anchoring to the query makes each iteration a contraction even when $\opT_\theta$ is mildly expansive, which provides a possible explanation for why truncated backpropagation works.
However, it also reveals that the iterates do \emph{not} converge to a fixed point of $\opT_\theta$, but to a biased solution.
We will see that this bias can be mitigated by scheduling the stepsize carefully.

Second, all of these guarantees hinge on controlling the Lipschitz constant of $\opT_\theta$.
In particular, for SwiGLU blocks \citep{shazeer2020}, it turns out that the gating makes the Lipschitz constant depend on a product of row and spectral norms, so the gate and value matrices should be constrained to the \emph{intersection} of the spectral and row-norm balls, which motivates a new optimizer.
It might sound like a small subset of matrices, but the value and gate matrices alone makes up 44\% of the non-embedding model parameters in a standard decoder-only Transformer.
These conclusions are not restricted to recurrent models, but inform how to train general deep models even without weight-tying.

\paragraph{Contributions}
Concretely, we make the following contributions:
\begin{enumerate}[label=(\roman*),leftmargin=2em]
    \item \textit{Halpern iteration for recurrent depth.}
    We formulate a set of reasonable desiderata for a recurrent forward pass, namely existence of (potentially multiple) fixed points with input-dependent selection and a truncated backpropagation window growing only polylogarithmically with the depth (\Cref{des:existence,des:selection,des:tbptt}).
    Halpern iteration, which we show is equivalent to input injection, falls out as one way of satisfying all three desiderata with a carefully scheduled anchor weight.
    Moreover, we show that Halpern iteration enjoys a robustness guarantee.
    \looseness=-1

    \item \textit{A Lipschitz neural parameterization.}
    Since these results rely on the Lipschitz condition of the neural parameterization $\opT_\theta$, we derive Lipschitz certificates for common blocks such as ReLU$^2$ based MLPs, SwiGLU, gated attention, and gated SSMs in \Cref{sec:lipschitz-parameterization}.
    This perspective reveals that SwiGLU and squared ReLU require a joint row-norm and spectral-norm constraint, while gates only require a row-norm constraint.
    As a side-effect we also find that pre-norm is essential specifically for SwiGLU/ReLU$^2$ and attention, and that normalization layers in general ensure existence of fixed points.

    \item \textit{\Sparrow{}: a drop-in replacement of Muon.}
    We propose \Sparrow{}, a parameter-group dependent optimizer that is tailored to these constraints (\Cref{sec:constrained-optimization}).
    It applies the spectral oracle at most once, so it has the same cost per step as Muon.
    Moreover, we derive a predetermined choice of constraint radii, so that no retuning is required.

    \item \textit{Experiments.}
    We show that the joint constraint is simultaneously more expressive and more robust than a spectral ball for SwiGLU blocks, and characterize when the two coincide (\Cref{sec:experiments-robustness}).
    Without any retuning or additional computational overhead, \Sparrow{} speeds up training of a small language model over Muon both with and without looping (\Cref{sec:experiments-speed}).
    We find that when combining \Sparrow{} with weight normalization as in MuonH one can ignore normalizing the update direction.
\end{enumerate}

\section{Related work}
\label{sec:related-work}

\paragraph{Fixed-point iterations}
The two classical schemes for finding fixed points date back to Krasnosel'ski\u\i--Mann (KM) iteration~\citep{mann1953}, which relaxes toward the current operator output, and Halpern iteration~\citep{halpern1967}, which repeatedly injects a fixed anchor.
The latter has received a lot of attention in recent years, due to the schedule $\beta_k=1/(k+2)$ achieving an accelerated rate~\citep{sabach2017,lieder2021}, which is worst-case optimal~\citep{park2022}.
Our work departs from this choice of stepsize and focuses on showing compatibility with truncated backpropagation with constant and logarithmic stepsize choices.
The closest work we are aware of is \citet{diakonikolas2025} who also study Halpern with fixed stepsize.

Classically, analysis of fixed point iterations relies on nonexpansiveness of the operator, see e.g, \citet{bauschke2017} for a standard treatment.
In recent years the community has moved beyond nonexpansiveness with structures such as conic (quasi-)nonexpansiveness \citep{bauschke2021,alacaoglu2024,pethick2023,pethick2025efficient}.
We instead follow \citet{diakonikolas2025} by allowing for mildly expansive operators in the case of fixed stepsize Halpern.
\looseness=-1

\paragraph{Implicit and recurrent-depth models}
Deep equilibrium models~\citep{bai2019deq,winston2020,sato2026} define the output as a fixed point of a weight-tied layer and compute gradients by implicit differentiation at that fixed point, which requires solving a linear system in the backward pass.
Recurrent-depth models~\citep{bansal2022,geiping2025} instead unroll the iteration explicitly and train with truncated backpropagation through the last iterations.
We directly adopt the input-injection structure of \citet{geiping2025}, which turns out to be an instance of Halpern iteration (\Cref{sec:halpern-iteration}).
Interestingly, in graph neural networks, APPNP~\citep{klicpera2019} also uses input injection, where it is motivated by personalization, which our Halpern connection provides a precise characterization of by showing that the fixed point is input-dependent.
The main difference from the recurrent-depth literature is that APPNP does not use truncated backpropagation.

\paragraph{Lipschitz neural networks}
Our approach of deriving weight constraints from the architecture is similar in spirit to the modular norm~\citep{large2024modular,bernstein2024modular}, and we extend this reasoning to squared ReLU, SwiGLU, gated attention and gated SSMs. %
We depart from the modular norm work by not focusing on smoothness, since smoothness with respect to the $\infty$-norm implies smoothness with respect to every $p$-norm with the same Lipschitz constant.
Instead, we take inspiration from the rates of \citet[Cor. 3.1]{pethick2026schatten}, where the leading term does not involve smoothness at all but depends on the initial distance to the solution $x^\star$ in the chosen norm $\|x^0-x^\star\|_p$, which we use to construct informative synthetic examples in \Cref{sec:experiments-robustness}.

We build on a long line of work on Lipschitz neural networks.
For MLPs, \citet{gouk2021} bound the Lipschitz constant by composing layerwise operator norms and enforce spectral constraints, as we do in \Cref{thm:nonexpansive-neural-block}.
A close work to our fixed-point perspective is \citet{combettes2020} which analyzes neural networks as averaged, and hence nonexpansive, operators. \citet{meunier2022} also consider nonexpansive layers by deriving them from convex potentials.
Regarding attention, dot-product attention is not globally Lipschitz~\citep{kim2021lipschitz} but was shown to be Lipschitz on bounded inputs~\citep{castin2024smooth}.
Our construction specifically relies on Sinkformers~\citep{sander2022}, which use a doubly stochastic attention matrix, so that mixing the tokens is nonexpansive.
We use this property to make the entire attention block Lipschitz, by additionally constraining the weights and normalizing the input.
Our arguments additionally rely on the Lipschitz properties of the normalization layers, specifically LayerNorm, which with gain $\gamma$ and stabilizer $\epsilon>0$ was shown to be $|\gamma|/\sqrt{\epsilon}$-Lipschitz in \citet{sato2026,emadi2026}.

Most existing certificates only constrain spectral norms.
Our certificates differ by also requiring row-norm constraints, either alone for gates or jointly with spectral constraints for SwiGLU.
We show that even simply changing the activation from ReLU to squared ReLU departs from the spectral norm as the appropriate weight constraint.

\paragraph{Row norms}
Row-wise normalization have recently been used extensively for deep learning.
\citet{loshchilov2025ngpt} normalizes the rows of the matrices reading from the residual stream in nGPT, The Scion framework~\citep{pethick2025scion} proposed a row-norm-based update, and NorMuon~\citep{li2025normuon}, Aurora~\citep{dewulf2026aurora} and Muown~\citep{lion2026muown} combine the spectral update of Muon with row-wise normalization or control.
\citet{lau2026symmetry} derive update directions from the symmetries of each layer, which for SwiGLU yielded a family of row-norm, one-sided spectral and hybrid updates for all three projections.
In contrast, our Lipschitz perspective determines which matrices should be constrained, and in which norms, specifically, jointly in row and spectral norm for the gate and value matrices, in spectral norm for the output matrix, and in row norm alone for attention gates.

\section{Problem and desiderata}
We start by formulating a set of reasonable desiderata for our recurrent procedure.
What we will find is that \emph{input injection} commonly used in recurrent models such as Huginn-0125 \citep{geiping2025} can satisfy these requirements.
Abstractly the recurrent architecture in \citet{geiping2025} takes the following form:
\begin{equation*}
y = \Bigl(
\underbrace{\opD}_{\text{decoder}}
\circ
\underbrace{\opG^k}_{\text{$k$ recurrent steps}}
\circ
\underbrace{\opE}_{\text{encoder}}
\Bigr)(x).
\end{equation*}
where $x$ is the input query, the encoder $\opE$ (also called the Prelude) maps it to an encoded query
$u=\opE(x)$, and the decoder $\opD$ (also called the Coda block) maps back the final latent state to the outputted answer $y$.
The notation $\opG^k$ denotes $k$ successive recurrent updates.

Our work focuses on the recurrence $\opG$ and asks how to choose this iterative process as a fixed point solver of some underlying operator.
Specifically, we are given a single-valued operator parameterized by $\theta$,
\[
\opT_\theta:\mathcal{H}\to\mathcal{H}
\]
where $\mathcal{H}$ is a finite-dimensional Hilbert space and $\opT_\theta$ has fixed-point set
\[
\cC_\theta := \Fix(\opT_\theta) := \{z\in\mathcal{H}:\opT_\theta(z)=z\}.
\]
Given an encoded query represented by $u\in\mathcal H$, we seek a
recurrent procedure
\begin{equation}\label{eq:recurrence}
z^{k+1}=\opG_{k,\theta,u}(z^k)
\end{equation}
which should be able to return a potentially distinct solution from $\cC_\theta$ based on each query $u$.
Furthermore, the parameters $\theta$ should be trainable with truncated backpropagation.
The following desiderata make these criteria precise.
\begin{desideratum}[Existence and multiple solutions]
\label{des:existence}
The fixed-point set $\cC_\theta$ is nonempty, and the operator class
permits $\cC_\theta$ to contain more than one point.
\end{desideratum}
If $\cC_\theta$ were a singleton then the underlying operator $\opT_\theta$ would not be rich enough to encode different solutions for different queries.
So we need a structural assumption on the operator that permits multiple solutions.
\begin{definition}
  The operator $\opT$ is $L$-Lipschitz if
  \begin{equation}
    \|\opT(x)-\opT(y)\|\le L\|x-y\|
    \qquad\text{for all }x,y\in\mathcal H.
    \label{eq:lipschitz}
  \end{equation}
\end{definition}
We call the operator \emph{contractive} if \eqref{eq:lipschitz} holds with some
$L<1$, \emph{nonexpansive} if it holds with $L=1$, and \emph{expansive}
if it holds only with $L>1$.  A contraction has a unique fixed point, so
we will require that the operator is at least nonexpansive.
See \citet{ryu2022algorithms} for an excellent overview.
\looseness=-1

\begin{desideratum}[Closest-point selection]
\label{des:selection}
For every input $u\in\mathcal H$ and initialization $z^0\in\mathcal H$,
the iterates generated by $z^{k+1}=\opG_{k,\theta,u}(z^k)$ satisfy
\[
z^k\longrightarrow\opP_{\cC_\theta}(u)
:=\arg\min_{z\in\cC_\theta}\tfrac12\|z-u\|^2.
\]
\end{desideratum}

\begin{figure}[t]
  \centering
  \includegraphics[width=0.6\linewidth]{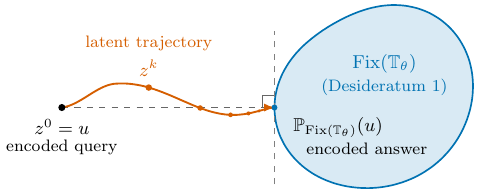}
  \caption{Closest-point selection (\Cref{des:selection}).
  The fixed-point set $\cC_\theta=\Fix(\opT_\theta)$ of a nonexpansive
  operator is closed and convex and may contain many solutions.  Started
  from the query/anchor $u$, the iterates $z^k$ converge to the closest fixed point
  $\opP_{\cC_\theta}(u)$.}
  \label{fig:closest-point}
\end{figure}

Through this desideratum, different queries can select different solutions from the
fixed-point set of one shared operator $\opT_\theta$
(\Cref{fig:closest-point}). 
The specific choice of the \emph{closest-point} is a natural choice since it also provides robustness guarantees as we will return to in \Cref{sec:query-robustness}.

Finally, we would like the recurrent procedure to be efficient to train, in the sense that we would \emph{not} want to backpropagate through all $K$ steps of the recurrence.
Instead, let $(\partial_\theta z^K)_{\mathrm{TBPTT}(H)}$ be obtained by
differentiating only the last $H$ updates with $z^{K-H}$ held constant.
We want that this approximation is not too far from the true gradient obtained by full backpropagation even when $H$ only very loosely depends on $K$.
\begin{desideratum}[Polylogarithmic-window TBPTT]
\label{des:tbptt}
For a differentiable recurrence \eqref{eq:recurrence} with $u$ and $z^0$ independent of
$\theta$, a backward window of size
$H=\operatorname{polylog}(K,1/\varepsilon)$ suffices to ensure
\[
\left\|\partial_\theta z^K
-(\partial_\theta z^K)_{\mathrm{TBPTT}(H)}\right\|
\le\varepsilon.
\]
\end{desideratum}
Here the norm is the induced operator norm, and $\operatorname{polylog}$
means at most polynomial growth in $\log K$ and $\log(1/\varepsilon)$.  
The window is capped at $K$, so shorter depths use full backpropagation.

\section{Halpern iteration for recurrent depth}
\label{sec:halpern-iteration}
We seek an iteration that satisfies \Cref{des:existence,des:selection,des:tbptt}.
It turns out that the classical method of Halpern iteration~\citep{halpern1967} provides a way to meet these requirements and that Halpern iteration in turn is exactly what is known as \emph{input injection}.
\paragraph{Halpern iteration}
For an anchor $u\in\mathcal H$ and a schedule $\beta_k\in[0,1]$,
Halpern iteration is
\equationbox{%
\begin{equation}
z^{k+1}
=
\opH_{\beta_k,u}(z^k)
:=
\beta_k u+(1-\beta_k)\opT_\theta(z^k).
\tag{Halpern}
\label{eq:halpern-iteration}
\end{equation}
}
The update is celebrated, since with the standard schedule $\beta_k=1/(k+2)$ and $z^0=u$, Halpern
iteration achieves an accelerated $O(1/k)$ fixed-point residual rate
for nonexpansive $\opT_\theta$ \citep{sabach2017,lieder2021}, which is worst-case optimal \citep{park2022}.
In this work we are interested in \ref{eq:halpern-iteration} for different reasons.
In particular, since we seek compatibility with truncated backpropagation, we instead consider constant and logarithmic stepsize schedules.

Let us first remark on where this update comes from to get a better understanding.

\paragraph{Fixed-step Halpern evaluates a resolvent}
\label{par:halpern-resolvent}
For a fixed schedule $\beta_k\equiv\beta\in(0,1)$, let
$\lambda:=(1-\beta)/\beta$ and define the displacement map
$\opD_\theta:=\opI-\opT_\theta$. For $z_\beta \in \Fix(\opH_{\beta,u})$, we have,
\[
z_\beta=\opH_{\beta,u}(z_\beta)
\quad\Longleftrightarrow\quad
z_\beta+\lambda\opD_\theta(z_\beta)=u
\quad\Longleftrightarrow\quad
z_\beta=\opJ_\lambda (u),
\]
with the resolvent
$\opJ_\lambda:=(\opI+\lambda\opD_\theta)^{-1}$.  
So the limit of \ref{eq:halpern-iteration} with fixed $\beta_k\equiv\beta$ evaluates one application of the resolvent.
This relationship already illuminates why Halpern is such a useful iteration for us.
For nonexpansive
$\opT_\theta$, $\opJ_\lambda (u)\to\opP_{\Fix(\opT_\theta)}(u)$ as
$\lambda\to\infty$~\citep{bauschke2017}, which is precisely \Cref{des:selection}.

\paragraph{Additive input injection is fixed-step Halpern}
It turns out that practitioners in deep learning have converged to the same update rule. 
Recurrent architectures repeatedly inject the original input into the
latent state, as in \citet{bansal2022,geiping2025}.
For additive injection, write the learned recurrent block as $\opS_\theta$:
\[
x^{k+1}=\opS_\theta(u+x^k),
\qquad x^0\sim\mathcal N(0,\sigma_0^2I).
\]
Define $z^k:=(u+x^k)/2$ and $\opT_\theta(z):=\opS_\theta(2z)$.
Then
\[
z^{k+1}
=\tfrac12u+\tfrac12\opS_\theta(2z^k)
=\tfrac12u+\tfrac12\opT_\theta(z^k)
=\opH_{1/2,u}(z^k).
\]
Thus additive input injection is exactly \ref{eq:halpern-iteration}
with $\beta_k\equiv1/2$ in the transformed state $z^k$ and
initialization $z^0=(u+x^0)/2$.

To cast light on input injection we now study Halpern iteration with constant and vanishing stepsize (\textit{c.f.} \Cref{tab:halpern-comparison}).
Proofs of the results in this section can be found in \Cref{app:halpern-iteration}.

\begin{table}[!t]
\centering
\small
\begin{tabular}{@{}p{0.22\linewidth}p{0.27\linewidth}p{0.24\linewidth}p{0.17\linewidth}@{}}
\toprule
Method & Converges to $\Fix(\opT_\theta)$? & Backward horizon & Operator\\
\midrule
Fixed-$\beta$ Halpern
    & No$^{\dagger}$
    & Depth-independent
    & $\operatorname{Lip}<\frac{1}{1-\beta}$\\
Logarithmic Halpern
    & Yes
    & Polylogarithmic in $K$
    & Nonexpansive\\
\bottomrule
\end{tabular}
\par\smallskip
\begin{minipage}{0.9\linewidth}
\footnotesize
$^{\dagger}$ Unless trivially $u\in\Fix(\opT_\theta)$.  %
\end{minipage}
\caption{Fixed-point selection, truncated backpropagation horizons and admissible operators of the two Halpern updates.}
\label{tab:halpern-comparison}
\end{table}

\subsection{Fixed-step Halpern: two out of three}

One recent work that also studies Halpern with a fixed stepsize is \citet{diakonikolas2025} which uses it to develop a restart strategy.
Here we study the implications of fixed stepsize Halpern for recurrent models and truncated backpropagation.

The primary reason why Halpern with fixed stepsize is so useful is that
the anchoring makes the update a contraction even when the underlying operator is mildly expansive.
This in turn is what enables truncated backpropagation as the following theorem makes precise.
\begin{proposition}[Fixed-step Halpern]
\label{prop:fixed-halpern}
Consider \ref{eq:halpern-iteration} with
$\beta_k\equiv\beta\in(0,1)$, and suppose $\opT_\theta$ is $L$-Lipschitz
with
\[
q:=(1-\beta)L<1.
\]
Note that the operator may be expansive ($L>1$) provided $L<1/(1-\beta)$.  Then:
\begin{enumerate}[label=(\roman*),leftmargin=2em]
    \item \emph{Contraction.} The update $\opH_{\beta,u}$ is a contraction with modulus $q$,
    so it has a unique fixed point $z_\beta$, and
    $\|z^k-z_\beta\|\le q^k\|z^0-z_\beta\|$ for every initialization.

    \item \emph{Depth-independent TBPTT window
    (\Cref{des:tbptt}).} If the iteration is differentiable
    along its trajectory, $u$ and $z^0$ are independent of $\theta$,
    and $\|\partial_\theta \opH_{\beta,u}(z^k)\|\le B$ uniformly in $k$,
    then, for $0\le H\le K$,
    \[
    \left\|
    \partial_\theta z^K-
    (\partial_\theta z^K)_{\mathrm{TBPTT}(H)}
    \right\|
    \le \tfrac{B}{1-q}q^H.
    \]
    Thus, a window
    $H=O(\log(1/\varepsilon))$, capped at $K$, suffices for error
    at most $\varepsilon$.

    \item \emph{Approximate fixed-point selection.} The selected equilibrium
    $z_\beta$ has fixed-point residual
    \[
    \|z_\beta-\opT_\theta(z_\beta)\|
    =\tfrac{\beta}{1-\beta}\|u-z_\beta\|.
    \]
    If $\opT_\theta$ is nonexpansive with nonempty
    $\cC_\theta=\Fix(\opT_\theta)$, this residual is at most
    $\tfrac{\beta}{1-\beta}\operatorname{dist}(u,\cC_\theta)$.
    It also retains the projection's variational inequality,
    $\langle u-z_\beta,c-z_\beta\rangle\le0$ for every $c\in\cC_\theta$.
    It is an exact fixed point if and only if $u\in\Fix(\opT_\theta)$.
\end{enumerate}
\end{proposition}
\paragraph{Tradeoff} There is a clear tradeoff that becomes apparent from this result.
\Cref{prop:fixed-halpern} needs only
$\operatorname{Lip}(\opT_\theta)<1/(1-\beta)$, e.g., $L<2$ at
$\beta=1/2$.  A larger anchor weight thus admits a more expansive
operator and a shorter TBPTT window, at the price of a larger solution bias.
Fixed-step Halpern can therefore satisfy
\Cref{des:existence,des:tbptt}, but it fails \Cref{des:selection}, since the iterates converge to a biased solution.
Next we resolve this issue by decaying $\beta_k$.

\subsection{Slowly vanishing Halpern}
\label{sec:slowly-vanishing-halpern}
Classical Halpern iteration removes the convergence bias by allowing the
anchor weight to vanish.  
However, the usual $\beta_k=1/(k+2)$ is too aggressive to be compatible with TBPTT.
We adopt a logarithmic schedule below, which makes the backpropagation window grow only polylogarithmically with the forward depth.
Here, $\log$ denotes the natural logarithm.
\begin{proposition}[Logarithmic-step Halpern]
\label{prop:log-halpern}
Assume that $\opT_\theta$ is nonexpansive with nonempty
$\cC_\theta=\Fix(\opT_\theta)$ (\Cref{des:existence}).  Then
\ref{eq:halpern-iteration} with $\beta_k:=\tfrac{1}{\log(k+3)}$
has the following properties.
\begin{enumerate}[label=(\roman*),leftmargin=2em]
    \item \emph{Closest-point selection
    (\Cref{des:selection}).} For every $z^0\in\mathcal H$,
    $
    z^k\longrightarrow \opP_{\cC_\theta}(u).
    $

    \item \emph{Polylogarithmic-window TBPTT
    (\Cref{des:tbptt}).} Suppose the recurrence is
    differentiable along its trajectory,
    $u$ and $z^0$ are independent of $\theta$, and
    $\sup_k\|\partial_\theta \opT_\theta(z^k)\|\le M$.
    Then, for $0\le H<K$,
    \begin{equation}
    \left\|
    \partial_\theta z^K-
    (\partial_\theta z^K)_{\mathrm{TBPTT}(H)}
    \right\|
    \le
    M\log(K+2)
    \exp\!\left(-\tfrac{H}{\log(K+2)}\right).
    \label{eq:log-halpern-tbptt}
    \end{equation}
    Thus, a window
    $H=O(\log K\,[\log\log K+\log(1/\varepsilon)])$
    suffices for absolute error at most $\varepsilon$.
\end{enumerate}
\end{proposition}
\paragraph{Tradeoff}
Thus, slowly vanishing Halpern can satisfy all of \Cref{des:existence,des:selection,des:tbptt}, with the price being that the operator $\opT_\theta$ is restricted to be nonexpansive.

\subsection{Query robustness}
\label{sec:query-robustness}
As required by \Cref{des:selection},
slowly vanishing Halpern selects the closest fixed point $\opP_{\cC_\theta}(u)$.
Since projection onto a closed convex set is (firmly) nonexpansive,
$\|\opP_{\cC_\theta}(u)-\opP_{\cC_\theta}(\widetilde u)\|\le\|u-\widetilde u\|$.
This provides a robustness guarantee stating that perturbing the query $u$ will only perturb the solution by as much.
We can say something stronger for the specific case of \ref{eq:halpern-iteration}.
The following theorem extends this robustness to every finite iterate and every schedule $\beta_k$.
\begin{theorem}[Query robustness of \ref{eq:halpern-iteration}]
\label{thm:halpern-robustness}
Let $\opT_\theta$ be $L$-Lipschitz, $\beta_k\in[0,1]$,
$\beta_{\min}:=\inf_k\beta_k > 1-\tfrac1L$, and
\begin{equation}
C:=1\ \text{ if }L\le1,
\qquad
C:=\tfrac{\beta_{\min}}{1-(1-\beta_{\min})L}\ \text{ if }L>1.
\label{eq:robustness-constant}
\end{equation}
If $(z^k)$ and $(\widetilde z^k)$ follow
$z^{k+1}=\opH_{\beta_k,u}(z^k)$ and
$\widetilde z^{k+1}=\opH_{\beta_k,\widetilde u}(\widetilde z^k)$,
and $\|u-\widetilde u\|,\|z^0-\widetilde z^0\|\le\varepsilon$,
then
\begin{equation}
\|z^k-\widetilde z^k\|\le C\varepsilon
\qquad\text{for every }k\ge0.
\label{eq:trajectory-robustness}
\end{equation}
\end{theorem}
The theorem applies even to expansive operators ($L>1$) for a possibly vanishing schedule as long as there is a floor $\beta_k\ge\beta_{\min}>1-1/L$.
This floor is necessary since the constant $C$ is sharp for $\beta_k\equiv \beta$, as realized by the simple example $\opT_\theta=L\opI$.

\section{A Lipschitz neural parameterization}
\label{sec:lipschitz-parameterization}
We have satisfied the stated desiderata, but we have assumed away one major obstacle, namely ensuring that the Lipschitz constant of the operator $\opT_\theta$ is controlled.
Specifically, \Cref{prop:fixed-halpern} requires $L < 1/(1-\beta)$.
This section is devoted to constructing such Lipschitz certificates along with ensuring existence of fixed points for common architectural components such as MLPs, SwiGLU, and attention.
We want to highlight that these results are not only relevant for recurrent models, but will inform us on how to train general deep models even without weight-tying.
Proofs and
supplementary results for this section are collected in
\Cref{app:neural-parameterization}.

\paragraph{Normalization layers}
Normalization layers can be made nonexpansive with appropriate parameterization.
For $x\in\mathbb{R}^d$, write $r:=\operatorname{rms}(x)=\|x\|/\sqrt d$.  For
$0<a<\tau$, we use
\begin{equation}
\opN_{\mathrm{RMS}}(x):=\frac{\tau x}{\sqrt{\tau^2+r^2}},
\qquad
\opN_{\tanh}(x):=
\frac{a+(\tau-a)\tanh\!\big(\frac{(r-a)_+}{\tau-a}\big)}
{\max\{r,a\}}\,x,
\label{eq:profiles}
\end{equation}
applied to each row of a sequence $X\in\mathbb{R}^{n\times d}$.
$\opN_{\mathrm{RMS}}$ is simply RMSNorm~\citep{zhang2019rmsnorm} with
scalar gain $g=\tau$ and stabilizer $\epsilon=\tau^2$.
For general $g>0$ and $\epsilon>0$, RMSNorm has Lipschitz constant
$g/\sqrt\epsilon$~\citep[Lemma~A.3]{chen2026hista}, so it is
nonexpansive when $g/\sqrt\epsilon\le1$.
We provide as an alternative, $\opN_{\tanh}$ (which is the identity up to RMS $a$ and smoothly saturates at RMS $\tau$) because its identity region permits multiple fixed points when composed with other nonexpansive operators, whereas compositions involving $\opN_{\mathrm{RMS}}$
have at most one.
Both $\opN_{\mathrm{RMS}}$ and $\opN_{\tanh}$ are instances of the more
general \emph{normalization layer} $\opN$ defined in
\Cref{def:normalization-profile} in \Cref{app:neural-parameterization} with
parameter $\omega=\tau/2$ for $\opN_{\mathrm{RMS}}$ and
$a\le\omega\le\tau$ for $\opN_{\tanh}$.
\looseness=-1

\paragraph{MLP}
As a warmup we start with a simple MLP using an $L_\sigma$-Lipschitz
activation function, such as ReLU for which $L_\sigma=1$.
\begin{theorem}[Lipschitz MLP block]
\label{thm:nonexpansive-neural-block}
Let $W_1:\mathbb{R}^d\to\mathbb{R}^m$ and
$W_2:\mathbb{R}^m\to\mathbb{R}^d$ be linear maps, let $b_1\in
\mathbb{R}^m$ and $b_2\in\mathbb{R}^d$, and suppose that
$\sigma:\mathbb{R}^m\to\mathbb{R}^m$ is $L_\sigma$-Lipschitz in the
Euclidean norm.  Define the raw MLP
\begin{equation}
\opF_{\mathrm{MLP}}(x)
:=W_2\sigma(W_1x+b_1)+b_2
\label{eq:raw-neural-M}
\end{equation}
and, for a normalization layer $\opN$, its input- and output-normalized
variants
\begin{equation}
\opM_{\mathrm{in}}
:=\opF_{\mathrm{MLP}}\circ\opN,
\qquad
\opM_{\mathrm{out}}
:=\opN\circ\opF_{\mathrm{MLP}}.
\label{eq:nonexpansive-neural-M}
\end{equation}
Then
\begin{equation}
\operatorname{Lip}(\opF_{\mathrm{MLP}}),\
\operatorname{Lip}(\opM_{\mathrm{in}}),\
\operatorname{Lip}(\opM_{\mathrm{out}})
\le\|W_2\|_{\mathrm{op}}L_\sigma\|W_1\|_{\mathrm{op}}.
\label{eq:mlp-lipschitz}
\end{equation}
\end{theorem}

\begin{remark}[Placement of the MLP normalization]
The bound \eqref{eq:mlp-lipschitz} already holds for the raw MLP, without
normalization.
However, the bounded range enforced by normalization layers is used for showing existence of fixed points (\Cref{app:fixed-point-existence}).
\end{remark}

\paragraph{SwiGLU and ReLU$^2$}
An activation function that leads to much more interesting conclusions is the SwiGLU block \citep{shazeer2020}, which has been widely adopted in the field.
Due to the gating mechanism there is an elementwise quadratic interaction of the input which in turns leads to a Jacobian that grows linearly in the input norm.
This will have two important consequences, one being that input normalization (also known as pre-norm) becomes essential, and second, that it naturally leads to a joint row-norm and spectral constraint on specific matrices.
Squared ReLU shares this structure and thus leads to a
similar bound involving both row and spectral norms, which additionally enjoys a particularly simple argument (see
\Cref{thm:nonexpansive-relu2}).

\begin{theorem}[Lipschitz SwiGLU block]
\label{thm:nonexpansive-swiglu}
Let $W_g,W_v:\mathbb{R}^d\to\mathbb{R}^m$ and
$W_o:\mathbb{R}^m\to\mathbb{R}^d$ be linear maps, and let $\opN$ be a
normalization layer.  Write
$\operatorname{SiLU}(t)=t/(1+e^{-t})$, applied coordinatewise, and let
$L_{\mathrm{SiLU}}:=\sup_t|\operatorname{SiLU}'(t)|\approx1.1$.  Define
the maximum row-norm $\|W\|_{2\to\infty}:=
\max_i\|W_{i,:}\|_2$ and the input-normalized SwiGLU map
\begin{equation}
\opM_{\mathrm{SwiGLU}}(x)
:=W_o\!\left[
\operatorname{SiLU}\!\left(W_g\opN(x)\right)
\odot W_v\opN(x)
\right].
\label{eq:swiglu-map}
\end{equation}
Then
\begin{equation}
\operatorname{Lip}(\opM_{\mathrm{SwiGLU}})
\le \sqrt d\,\omega\|W_o\|_{\mathrm{op}}
\left[
L_{\mathrm{SiLU}}\|W_v\|_{2\to\infty}\|W_g\|_{\mathrm{op}}
+\|W_g\|_{2\to\infty}\|W_v\|_{\mathrm{op}}
\right].
\label{eq:swiglu-lipschitz}
\end{equation}
\end{theorem}

\begin{remark}[Spectral and rownorm split for SwiGLU]
\label{rem:swiglu-budget}
There is freedom in how we allocate the Lipschitz budget.
If $\|W_o\|_{\mathrm{op}}\le1$ and $W_g,W_v$ have spectral norms at most $s$
and row norms at most $r$, then
$\operatorname{Lip}(\opM_{\mathrm{SwiGLU}})
\le(1+L_{\mathrm{SiLU}})\sqrt d\,\omega\,rs$.  The certificate depends on the
radii only through the budget $B:=rs$, which nonexpansiveness fixes at
$B\le1/((1+L_{\mathrm{SiLU}})\sqrt d\,\omega)$, with
the ratio $\kappa:=r/s\in(0,1]$ being free.
To choose a split, consider what we will call a \emph{saturated} matrix $W\in\mathbb R^{m\times d}$
with $m\ge d$, where all $d$ singular values equal to $s$, and all $m$ row norms
equal to $r$.  Then $mr^2=\|W\|_F^2=ds^2$, giving $r/s=\sqrt{d/m}$.
For a wide matrix, $m<d$, all $m$ singular values equal $s$, so $WW^\top=s^2I_m$ and the
rows are orthogonal with norm exactly $r=s$.
We therefore recommend $\kappa=\min\{1,\sqrt{d/m}\}$, as in
\Cref{alg:sparrow}, which makes the row and spectral bounds simultaneously
tight for this class of fully saturated matrices
(see \Cref{sec:experiments-robustness}).
\end{remark}

\begin{insightbox}
SwiGLU blocks and MLP blocks using ReLU$^2$ should optimize over the intersection of the spectral and row norm unit balls for which we propose the \Sparrow{} optimizer (\textit{c.f.} \Cref{sec:constrained-optimization}).
\end{insightbox}

\begin{table}[t]
  \centering
  \small
  \setlength{\tabcolsep}{3pt}
  \renewcommand{\arraystretch}{1.35}
  \begin{threeparttable}
  \begin{tabular*}{\linewidth}{@{}l@{\extracolsep{\fill}}llll@{}}
    \toprule
    \textbf{Block} & \textbf{Normalize}
      & \multicolumn{2}{c}{\textbf{Constrained matrices}}
      & \textbf{Lipschitz bound}\\
    \cmidrule(lr){3-4}
      & & Spectral ($2 \rightarrow 2$) & Row ($2 \rightarrow \infty$) & \\
    \midrule
    MLP {\scriptsize (\Cref{thm:nonexpansive-neural-block})}
      & None\tnote{\textdagger}
      & $W_1,W_2$
      & ---
      & $L_\sigma s_1s_2$\\
    $\operatorname{ReLU}^2$ MLP {\scriptsize (\Cref{thm:nonexpansive-relu2})}
      & Input
      & $W_1,W_2$
      & $W_1$
      & $2\sqrt d\,\omega s_1s_2r_1$\\
    SwiGLU {\scriptsize (\Cref{thm:nonexpansive-swiglu})}
      & Input
      & $W_o,W_g,W_v$
      & $W_g,W_v$
      & $\sqrt d\,\omega s_o
        (L_{\mathrm{SiLU}}r_vs_g+r_gs_v)$\\
    Attention {\scriptsize (\Cref{thm:nonexpansive-attention})}
      & Input
      & $W_Q,W_K,W_V,W_O$
      & ---
      & $(1+2Be^{4B})s_Vs_O$\\
    Gated attention {\scriptsize (\Cref{thm:nonexpansive-gated-attention})}
      & Input
      & $W_O$
      & $W_a$
      & $cs_O(L_U+Mr_a/4)$\\
    Gated SSM {\scriptsize (\Cref{thm:nonexpansive-gated-ssm})}
      & Input
      & $W_V,W_O$
      & $W_a$
      & $\frac{s_Os_V}{1-\rho}
        (1+\frac{\sqrt d\,\tau r_a}{4(1-\rho)})$\\
    \bottomrule
  \end{tabular*}
  \begin{tablenotes}
    \footnotesize
    \item[\textdagger] The MLP needs no normalization for its Lipschitz bound, but the
    boundedness-based existence guarantee relies on input or output
    normalization (\Cref{thm:bounded-normalization-fixed-point}).
  \end{tablenotes}
  \caption{Lipschitz bounds of the blocks.  
  Let
  $s_i:=\|W_i\|_{\mathrm{op}}$ and $r_i:=\|W_i\|_{2\to\infty}$.
  }
  \label{tab:nonexpansive-summary}
  \end{threeparttable}
\end{table}

\paragraph{Attention}
We use Sinkhorn-based attention~\citep{sander2022}, whose Lipschitz certificate is given in \Cref{thm:nonexpansive-attention} (\Cref{app:sinkhorn-attention}) revealing spectral matrix constraints.
Similarly to the SwiGLU block, input normalization turns out to be crucial for attention.
However, note that no QK-normalization is needed for the Lipschitz certificate.
The result can be extended to softmax attention by adapting
\citet[Theorem~3.5]{castin2024smooth}, where the importance of pre-norm is already apparent since the result relies on boundedness of the input.

\paragraph{Gated architectures}
In the appendix we additionally derive Lipschitz certificates for gated
attention and gated SSMs in \Cref{app:gated-attention,app:gated-ssm},
which naturally leads to a constraint on the row norms of the gate weights without requiring a spectral bound. 
Interestingly, in a recent hero run, Marin reported
attention-gate norm growth~\citep{marin2026gatenorm} and intervened by
adding decoupled weight decay to the attention-gate~\citep{marin2026gatedecay}, while using the spectral-based optimizer MuonH. 
Our theory suggests instead optimizing the gate matrices
with a row-norm-based optimizer, such as Scion ({\footnotesize RowNorm})~\citep{pethick2025scion}.
This is the attention-gate branch of \Sparrow{}, which
\Cref{sec:experiments-gate} shows helps when the gate matrix is tall.
\looseness=-1

\paragraph{Combining attention and MLP blocks}
Transformer architectures evaluate attention and MLP blocks either
serially or in parallel.  Suppressing normalization
and parameters, the serial block with residual connections as in
ResNet~\citep{he2016} and the parallel block
used by e.g., PaLM and PAF~\citep{chowdhery2022,sonkar2023} have the
forms
\begin{align*}
\text{serial residual:}\quad
&X^+=(\opI+\opM)\circ(\opI+\opA)(X),\\
\text{parallel residual:}\quad
&X^+=X+\opA(X)+\opM(X).
\end{align*}
Because of the unscaled identity residual, their Lipschitz constants are
up to $(1+L_M)(1+L_A)$ and $1+L_A+L_M$ when $\opA$ and $\opM$ are
$L_A$- and $L_M$-Lipschitz, so neither is nonexpansive even when the
branches are.  We instead replace the additions by convex combinations.

\begin{proposition}[Serial and parallel Lipschitz blocks]
\label{prop:nonexpansive-parallel-block}
Let $\opA_\theta,\opM_\theta:\mathcal H\to\mathcal H$ be $L_A$- and
$L_M$-Lipschitz, and let $\alpha_A,\alpha_M,\alpha\in[0,1]$.  Define
\begin{align}
\opF_A&:=(1-\alpha_A)\opI+\alpha_A\opA_\theta,
&
\opF_M&:=(1-\alpha_M)\opI+\alpha_M\opM_\theta,
\label{eq:serial-averaged-blocks}\\
\opT_{\mathrm{ser}}&:=\opF_M\circ\opF_A,
&
\opT_{\mathrm{par}}&:=(1-\alpha)\opA_\theta+\alpha\opM_\theta.
\label{eq:parallel-convex-block}
\end{align}
Then
$\quad
\operatorname{Lip}(\opT_{\mathrm{ser}})
\le(1-\alpha_M+\alpha_ML_M)(1-\alpha_A+\alpha_AL_A),
\quad\text{and}\quad
\operatorname{Lip}(\opT_{\mathrm{par}})
\le(1-\alpha)L_A+\alpha L_M.
$
\end{proposition}

\Cref{tab:nonexpansive-summary} summarizes our Lipschitz bounds for the blocks and their implied weight constraints.

\paragraph{Fixed-point existence}
Normalization layers are sufficient to ensure that a fixed point exists, since every block of the form $\opG\circ\opN\circ\opF$
with continuous $\opF,\opG$, which covers the (normalized) MLP, SwiGLU, ReLU$^2$, attention and gated blocks above, has a fixed point
regardless of its Lipschitz constant
(\Cref{thm:bounded-normalization-fixed-point} in
\Cref{app:fixed-point-existence}).

\section{Optimizing the constrained weights}
\label{sec:constrained-optimization}

Our Lipschitz certificates show that our weight matrices should be constrained in particular ways as summarized in \Cref{tab:nonexpansive-summary}. 
This section deals with how our optimizer takes these constraints into account.
In particular, we will focus on computing the linear maximization oracle (LMO) and projection:
\begin{equation}
\lmo_{\mathcal K}(G):=\operatorname*{arg\,max}_{Z\in\mathcal K}
\langle G,Z\rangle_F,
\qquad
\opP_{\mathcal K}(Y):=\operatorname*{arg\,min}_{Z\in\mathcal K}
\tfrac12\|Z-Y\|_F^2.
\label{eq:lmo-definition}
\end{equation}
Both minimize $q(W):=\tfrac a2\|W\|_F^2-\langle H,W\rangle_F$ over
$\mathcal K$, with $(a,H)=(0,G)$ for the LMO and $(a,H)=(1,Y)$ for the
projection.

\paragraph{Single-constraint oracles}
To set up notation, consider the spectral and row-norm oracles under the single constraints $\|W\|_{\mathrm{op}}\le s$ and
$\|W\|_{2\to\infty}\le r$, respectively.
For a compact SVD $X=U\operatorname{diag}(\sigma_j)V^\top$,
\[
\begin{aligned}
\text{LMO:}&&\mathsf{Spectral}_s(X)&=sUV^\top,
&\mathsf{Row}_r(X)_{i,:}&=r\tfrac{X_{i,:}}{\|X_{i,:}\|_2},\\
\text{projection:}&&\mathsf{SpectralClip}_s(X)&=U\operatorname{diag}(\min\{\sigma_j,s\})V^\top,
&\mathsf{RowClip}_r(X)_{i,:}&=\tfrac{X_{i,:}}{\max\{1,\|X_{i,:}\|_2/r\}},
\end{aligned}
\]

\newlength{\sparrowlabelwidth}
\begin{algorithm}[t]
\caption{\textsf{Sparrow} update direction.  It applies $\mathsf{Spectral}_s$ at most once and is always feasible.}
\label{alg:sparrow}
\begin{algorithmic}[1]
\renewcommand{\algorithmicensure}{\textbf{Return:}}
\Require gradient estimate $G\in\mathbb R^{m\times d}$, radii $s$ and $r=s\min\{1,\sqrt{d/m}\}$ (see \Cref{rem:swiglu-budget})
\Ensure depending on the (hidden) parameter group of $G$:
\settowidth{\sparrowlabelwidth}{SwiGLU gate and value matrices ($m>d$)\textsuperscript{\textdagger}:}
\State \makebox[\sparrowlabelwidth][l]{SwiGLU gate and value matrices ($m>d$)\textsuperscript{\textdagger}:}\quad $(\mathsf{RowClip}_r\circ\mathsf{Spectral}_s\circ\mathsf{Row}_r)(G)$
\State \makebox[\sparrowlabelwidth][l]{Attention-gate/SSM-gate matrices:}\quad $\mathsf{Row}_r(G)$
\State \makebox[\sparrowlabelwidth][l]{All other matrices:}\quad $\mathsf{Spectral}_s(G)$
\end{algorithmic}
{\footnotesize\textsuperscript{\textdagger}Recommended solver, see \Cref{sec:rowspectral-solver} for alternative solvers and \Cref{alg:sparrow-c} for the general case.\par}
\end{algorithm}

\subsection{Joint row-norm and spectral-norm constraints}
\label{sec:rowspectral-solver}

The constraint that requires extra care is the SwiGLU bound \eqref{eq:swiglu-lipschitz} constraining $W_g,W_v$ to
\begin{equation}
\mathcal K_{r,s}:=\left\{W\in\mathbb R^{m\times d}:
\|W\|_{2\to\infty}\le r,\ \|W\|_{\mathrm{op}}\le s\right\},
\qquad r,s>0.
\label{eq:row-spectral-feasible-set}
\end{equation}
Since $\|W\|_{2\to\infty}\le\|W\|_{\mathrm{op}}\le\sqrt m\,\|W\|_{2\to\infty}$,
both constraints are active only when $r<s<\sqrt m\,r$.
For $m\ge d$, the recommended split $r/s=\sqrt{d/m}$ in
\Cref{rem:swiglu-budget} makes both constraints nonredundant when
$m>d>1$.  At the endpoints $m=d$ and $d=1$,
the spectral bound implies the row bound and the row bound implies
the spectral bound, respectively.
In other words, for square matrices the row bound is redundant.

\paragraph{Sparrow}
Since every method needs a name, we will call an (approximate) solution to the LMO and projection oracles
\Sparrow{} which is short for \emph{\underline{Sp}ectral \underline{a}nd
\underline{r}ow-norm \underline{r}estricted \underline{o}ptimization of
\underline{w}eights}.  \Cref{alg:sparrow} gives the LMO update we recommend based on the weight constraints summarized in \Cref{tab:nonexpansive-summary} and the approximation of the joint constraint developed below.
In practice, the LMO can be combined with standard techniques such as a Nesterov momentum based gradient estimator, weight decay and weight normalization (\Cref{app:lmo-optimizer}) as done in our accompanying PyTorch and Jax implementation.
\Sparrow{} provides a drop-in replacement of Muon/MuonH that leads to improvements without any retuning.
\looseness=-1

\paragraph{Coarse feasible approximation}
Neither oracle of $\mathcal K_{r,s}$ has a closed form.  A cheap candidate
applies the oracle of one (or two) constraints and then clips with respect to the other constraint:
\begin{equation}
\begin{aligned}
\text{LMO:}\quad
&U_1=(\mathsf{RowClip}_r\circ\mathsf{Spectral}_s)(G),\quad
U_2=(\mathsf{SpectralClip}_s\circ\mathsf{Row}_r)(G),\\
&U_3=(\mathsf{RowClip}_r\circ\mathsf{Spectral}_s\circ\mathsf{Row}_r)(G),\\
\text{projection:}\quad
&V_1=(\mathsf{RowClip}_r\circ\mathsf{SpectralClip}_s)(Y),\quad
V_2=(\mathsf{SpectralClip}_s\circ\mathsf{RowClip}_r)(Y).
\end{aligned}
\label{eq:rowclip-polar-approximation}
\end{equation}
Alternatively, one can keep the candidate with the
smallest value of $q$, which costs $O(md)$ to check.
Importantly, the output is feasible, since row clipping scales rows by factors at most
one, so it cannot increase the spectral norm, and spectral clipping cannot
increase any row norm. 
In contrast, Aurora is only feasible for the spectral norm and not necessarily for the row norm.
\looseness=-1

As we shall see next, these coarse approximations can be seen as performing one step of an iterate splitting algorithm for solving the dual problem of \eqref{eq:lmo-definition} with $\mathcal{K}=\mathcal{K}_{r,s}$.

\paragraph{Dual problem and forward-backward splitting}
Since $\mathcal K_{r,s}$ is the intersection of the balls
$\{\|Z\|_{2\to\infty}\le r\}$ and $\{\|Z\|_{\mathrm{op}}\le s\}$, whose support
functions are $r\|\cdot\|_{2,1}$ and $s\|\cdot\|_{\mathrm{nuc}}$ with
$\|A\|_{2,1}:=\sum_i\|A_{i,:}\|_2$, the LMO has the dual
\begin{equation}
\max_{Z\in\mathcal K_{r,s}}\langle G,Z\rangle
=\min_{A+B=G}\ r\|A\|_{2,1}+s\|B\|_{\mathrm{nuc}}.
\label{eq:lmo-dual}
\end{equation}
Forward-backward splitting on \eqref{eq:lmo-dual}, reduces to taking a gradient step on
$s\|G-A\|_{\mathrm{nuc}}$ followed by a proximal step on $r\|A\|_{2,1}$, which reads
\begin{equation}
P:=\mathsf{Spectral}_s(G-A),\qquad
A\leftarrow\mathsf{RowShrink}_{\gamma r}(A+\gamma P),\qquad
Z:=\mathsf{RowClip}_r(P),
\tag{FB}
\label{eq:fb}
\end{equation}
where $\mathsf{RowShrink}_t:=I-\mathsf{RowClip}_t$ is the proximal map of
$t\|\cdot\|_{2,1}$ (Moreau's decomposition).
An intelligent initialization is $A=\mathsf{RowShrink}_c(G)$, since this is the minimizer of
\eqref{eq:lmo-dual} with $s\|B\|_{\mathrm{nuc}}$ replaced by the quadratic
$\tfrac{r}{2c}\|B\|_F^2$.
Started from
$A=\mathsf{RowShrink}_c(G)$, so that $G-A=\mathsf{RowClip}_c(G)$, the first
output is
\[
Z=(\mathsf{RowClip}_r\circ\mathsf{Spectral}_s\circ\mathsf{RowClip}_c)(G).
\]
This is exactly our coarse approximations $U_1$ for $c\to\infty$ (indeed for $c\ge\max_i\|G_{i,:}\|_2$) and
$U_3$ for $c\to0$ (indeed for $c\le\min_i\|G_{i,:}\|_2$, where every row of
$\mathsf{RowClip}_c(G)$ has norm $c$ and $\mathsf{Spectral}_s$ is invariant to
positive scaling). 
See \Cref{app:sparrow-c} for a numerically stable implementation.

The choice of the splitting is in some sense arbitrary.
We could just as well exchange the roles of the two terms in \eqref{eq:lmo-dual} by taking a gradient step
on $r\|G-B\|_{2,1}$ and a proximal step on $s\|B\|_{\mathrm{nuc}}$ with output
$\mathsf{SpectralClip}_s(\mathsf{Row}_r(G-B))$, which gives $U_2$ with $B=0$.
The main disadvantage is computational since $\mathsf{SpectralClip}_s$ can be more expensive as we touch on later.

\paragraph{Iterative refinement through operator splitting}
If the coarse one step approximations are not accurate enough, one could iterate \ref{eq:fb}, but since
$\mathsf{Spectral}_s$ is not Lipschitz, \ref{eq:fb} converges only locally and needs a
small step size.  Douglas-Rachford splitting \eqref{eq:drs}, or equivalently ADMM on
the dual \eqref{eq:lmo-dual}, solves the joint problem for every step size, at the cost of one
$\mathsf{SpectralClip}_s$ per step.
It splits the objective into
$f:=q+\iota_{\{\|W\|_{2\to\infty}\le r\}}$ and
$g:=\iota_{\{\|W\|_{\mathrm{op}}\le s\}}$, where $\iota$ is the convex
indicator, with proximal maps
\[
\operatorname{prox}_{\gamma f}(Z)
=\mathsf{RowClip}_r\!\left(\tfrac{Z+\gamma H}{1+\gamma a}\right),
\qquad
\operatorname{prox}_{\gamma g}(Z)=\mathsf{SpectralClip}_s(Z).
\]
Since the spectral step is typically inexact while the row step is exact, one can use
inexact DRS (iDRS)~\citep{pethick2025ifeddr}, which allows for a relative error condition on the inexactness and provides a stopping criterion for the inner solver.
For the projection, alternating projection (POCS) of the two clipping
maps is another option, for which one step yields $V_1,V_2$ in \eqref{eq:rowclip-polar-approximation}.

We compare these approaches with inexact, low-rank spectral steps in \Cref{app:lmo-idrs}.
In practice, however, Newton-Schulz iterations~\citep{higham2008functions}
and PolarExpress~\citep{amsel2025polar} are typically used to approximate  $\mathsf{Spectral}_s$
using only matrix products.  $\mathsf{SpectralClip}_s$ can similarly be approximated through two approximations of $\mathsf{Spectral}_s$ (\textit{c.f.} \Cref{app:matmul-spectral}).

\section{Experiments}
\label{sec:experiments}

\subsection{Robustness-accuracy pareto frontier}
\label{sec:experiments-robustness}

We create a synthetic example where we can clearly investigate when (and when not) our derived constraints can be beneficial over a purely spectrally constrained based approach like Muon.
Specifically, inspired by the rates of~\citet[Cor.~3.1]{pethick2026schatten}, which
depend on the distance to solution $\|x^0-x^\star\|$ in some chosen geometry, we construct a teacher that is close in the joint row
and spectral geometry, while simultaneously being far (or close) in other geometries.

\paragraph{Setup}
We train the SwiGLU block $\opM(x)=\opG(\opN_{\mathrm{RMS}}(x))$ with
$\opG(z)=W_o[\operatorname{SiLU}(W_gz)\odot W_vz]$,
$\|W_o\|_{\mathrm{op}}\le1$ and $W_g,W_v\in\mathbb R^{m\times d}$.
Since the certificate depends on $W_g,W_v$ only through the budget $B=rs$,
the product of their maximal row norm $r$ and spectral norm $s$
(\Cref{rem:swiglu-budget}), we compare the spectral ball
$\{\|W\|_{\mathrm{op}}\le\sqrt B\}$ with the joint set
$\mathcal K_{\sqrt{B\kappa},\sqrt{B/\kappa}}$ at the same budget $B$, where
$\kappa=r/s$ is the split. %

The task is ten-class classification with labels given by the argmax of a
teacher with $d=64$ and $m=4d$, which is roughly nonexpansive (\Cref{app:classification-details}).
We describe a teacher by the ratio $\kappa^\star$ of its maximal row norm to
its spectral norm, and consider a Gaussian teacher whose mass is \emph{spread} across the rows
($\kappa^\star=0.43$) and a \emph{concentrated} teacher with only 8 nonzero rows in $W_g$ and $W_v$ ($\kappa^\star=0.90$).

We compare SGD with a Frobenius ball, Adam with an entrywise box, Muon with a
spectral ball, and our update with the joint set constraint (using the approximations $U_1$ and $V_1$ from \eqref{eq:rowclip-polar-approximation} for the LMO and projection, respectively), all with the same budget $B$.
Robustness is measured by the certified radius of each test point,
computed from the Lipschitz certificate of the trained weights.
For each method $M$ we sweep over step sizes and budgets for three seeds and
report the Pareto frontier $R_M(a)$ defined as the largest median radius among runs with
clean accuracy at least $a$ (see \Cref{app:experimental-details} for details).

\begin{figure}[t]
    \centering
    \includegraphics[width=\linewidth]{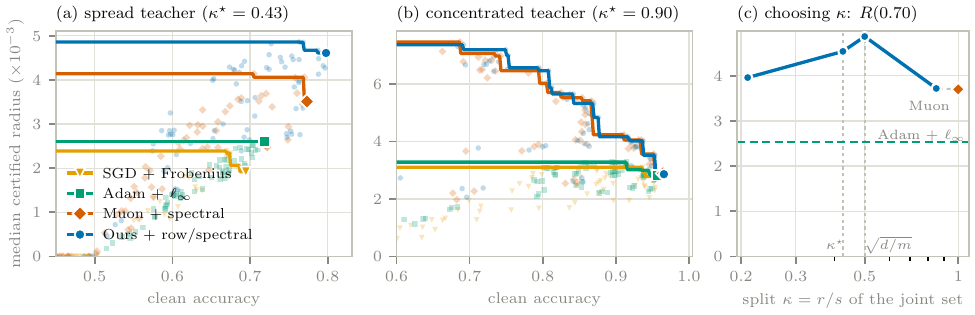}
    \caption{Robustness.  Row/spectral constraints certify larger radii for a given
    accuracy.  (a, b)~Each point is one run and each line is the frontier
    $R_M(a)$ (3 seeds). (c) We can set the budget split as $\kappa=\sqrt{d/m}$ without needing to know the ground truth $\kappa^*$.} %
    \label{fig:ts-robust}
\end{figure}

\paragraph{Spectral constraint}
We observe in \Cref{fig:ts-robust} that the spectral constraint certifies much
larger radii than the Frobenius and entrywise constraints.
At $70\%$ accuracy on the spread teacher, the radius certified by Muon is
$1.6\times$ that of Adam, and SGD and Adam
require $64$ to $4096$ times the budget of the teacher to fit the task.
The budgets of the four constraints are not directly comparable, since their worst case bound can be far from a typical case. 
However, the comparison remains fair, since compute every radius from the trained weights instead.

\paragraph{The row norm constraint}
On the spread teacher, the joint set dominates Muon at every accuracy level,
with $1.2\times$ the radius at $70\%$ accuracy and a higher maximal accuracy
($0.793$ vs.\ $0.771$).
The same holds with $3\times$ longer training (cf.\ \Cref{tab:ts-robust}).
The row constraint helps because the spectral ball cannot make use of the
width as we show in \Cref{app:classification-details} by making $W_g$ and
$W_v$ increasingly tall.
On the concentrated teacher, for which $\kappa^\star$ is close to one, the two
methods have a matching pareto front.
In other words, the joint constraint is only useful if the problem truly requires all the parameters.

\paragraph{Choosing the radius ratio}
Since $\kappa^\star$ is not known in practice, we also vary the split $\kappa$
of the joint set on the spread teacher.
\Cref{fig:ts-robust}c shows that $R(0.70)$ peaks near the teacher-free split
$\kappa=\sqrt{d/m}$ of \Cref{rem:swiglu-budget}, which is used in \Sparrow{}.
We observe that on the concentrated teacher this choice also does not hurt. %

\subsection{\Sparrow{} as a drop-in replacement for Muon}
\label{sec:experiments-speed}

Without the projection, \Sparrow{} can be used as an update direction for
unconstrained weights in place of Muon.
We train the character-level Shakespeare model of nanoGPT (4 layers, $d=128$,
RMSNorm, SwiGLU width $m=4d$) for one epoch.
The hidden matrices are updated as $W\leftarrow W-\eta\lmo(M)$, where $M$ is
the momentum, and the embeddings and head are trained with AdamW.
Muon uses $\lmo=\mathsf{Spectral}_1$ with its best step size $\eta=0.01$.
Aurora \citep{dewulf2026aurora} and \Sparrow{} only change the $\lmo$ for
$W_g$ and $W_v$ and otherwise use the same configuration as Muon (see \Cref{app:speed-details}).

\begin{figure}[t]
    \centering
    \includegraphics[width=\linewidth]{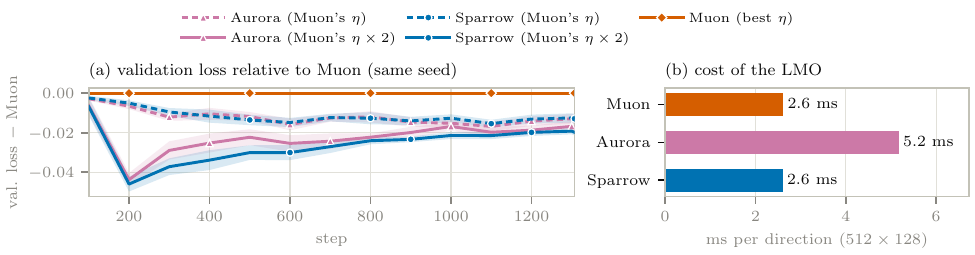}
    \caption{\textsf{Sparrow} lowers the validation loss at no additional cost per step.  (a)~Validation
    loss minus Muon's at the same step and seed (mean over 10 seeds with one standard error), with Muon at its best step size $\eta$ and
    Aurora and \textsf{Sparrow} at $1\times$ and $2\times$ the step size of Muon.  (b)~The LMO in \Sparrow{} has no wall-clock time overhead over Muon in contrast with Aurora.}
    \label{fig:speed}
\end{figure}

\begin{figure}[t]
    \centering
    \includegraphics[width=\linewidth]{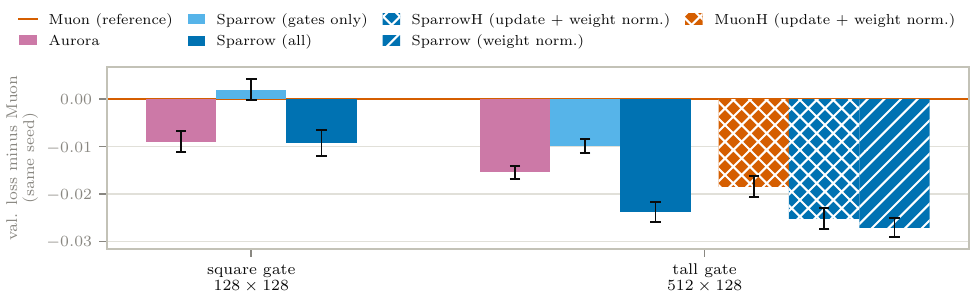}
    \caption{Gated attention: final validation loss minus that of Muon on
    the same seed (mean and standard error over ten seeds).
    The row-norm update of the gate in \Sparrow{} helps only when the gate is tall.
    When combined with Frobenius normalization as in MuonH, one can ignore normalizing the update direction in \Sparrow{} without sacrificing performance.}
    \label{fig:gate}
\end{figure}

\paragraph{Speedup}
\Cref{fig:speed}a shows the validation loss of each method minus that of Muon
on the same seed.  At Muon's step size, \Sparrow{} lowers the final validation loss on all 10
seeds, by $0.013\pm0.002$ on average, and reaches the final loss of Muon
after $1220$ instead of $1302$ steps ($1.07\times$).
At twice that step size, it lowers the loss by $0.020\pm0.002$ ($1.11\times$), on par with Aurora at
the same step size ($0.017\pm0.003$, $1.09\times$).
A direction of \Sparrow{} takes the same $2.6$\,ms as Muon's, since the additional row normalization and
row clipping are essentially free in comparison with the spectral solver, whereas Aurora's two polar
factors double the cost (\Cref{fig:speed}b).
Additionally, in contrast with Aurora, the update direction of \Sparrow{} lies
in $\mathcal K_{r,s}$ (\Cref{app:speed-details} compares both methods with the exact LMO).

\paragraph{Gated attention}
\label{sec:experiments-gate}
The Lipschitz constant of the gate depends only on $\|W_a\|_{2\to\infty}$
(\Cref{thm:nonexpansive-gated-attention}), so \Sparrow{} updates $W_a$ with
$\mathsf{Row}_r$ and drops the spectral constraint on the gate altogether.
This only makes a difference when $W_a$ is tall, since otherwise every row of
$\mathsf{Spectral}_s(G)$ already has norm $s$.
Tall gates are indeed used in practice, e.g., Qwen3-Next gates $4096$ channels at
width $2048$~\citep{qwen2025qwen3next,qiu2025gated}.
We add a gate to the model above with 4 heads of
width 32, which gives a square $128\times128$ gate, or 128, which gives a tall
$512\times128$ gate.
We compare Muon, Aurora on every tall matrix as recommended, \Sparrow{} on the gates only,
and \Sparrow{} on all matrices, each over ten seeds.

\begin{itemize}[leftmargin=2em]
    \item \emph{Square gate.}  Using \Sparrow{} on the gate alone has no
    effect ($+0.002\pm0.002$), as expected (\Cref{fig:gate}), so \Sparrow{} is safe to use even when the gate matrices are square. 
    The ablation also tells us that the large improvement from \Sparrow{} (all) comes exclusively from applying \Sparrow{} to the attention gates.
    \item \emph{Tall gate.}  
    Applying Aurora improves the most over Muon in the tall gate setting ($0.015\pm0.001$).
    However, dropping the spectral constraint entirely as in \Sparrow{} leads to a more than $1.5\times$ improvement compared with Aurora ($0.024\pm0.002$), suggesting that the row-norm is the more appropriate constraint for the attention gates, and that Aurora should \emph{not} be applied to these matrices despite them being tall.
    \item \emph{Weight normalization.}  MuonH~\citep{wen2026hyperball} keeps
    each hidden matrix at its initial Frobenius norm and additionally normalizes the update direction, which lowers the loss with
    the tall gate by $0.018\pm0.002$.\footnote{The improvement of MuonH might be partially caused by changing away from zero initialization, which is otherwise incompatible with MuonH.} 
    Combining \Sparrow{} with weight normalization the loss is further improved by $0.007$--$0.009$ ($\pm0.002$) even without using update direction normalization, which suggests that the two are
    complementary.
\end{itemize}

\begin{figure}[t]
    \centering
    \includegraphics[width=\linewidth]{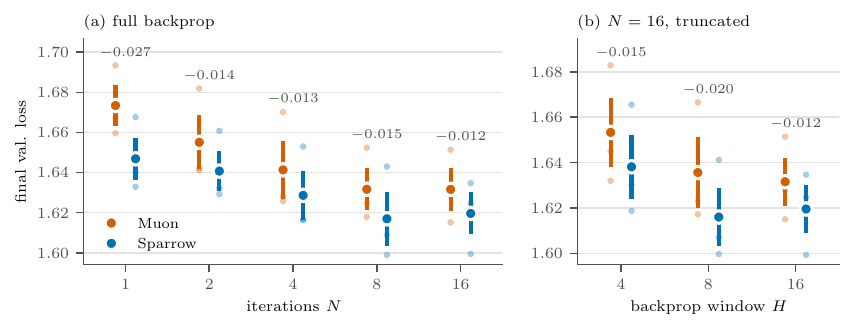}
    \caption{Looped model. Final validation loss at each method's best step size (mean $\pm$ one standard error
    over three seeds, the numbers indicate \Sparrow{} minus Muon on the same seed).
    \Sparrow{} consistently outperform Muon without retuning.}
    \label{fig:loop-shakespeare}
\end{figure}

\paragraph{Looped models}
\label{sec:experiments-looped}
We extend the setup with the tall gate to a looped model with a prelude layer, a core layer iterated $N$
times with fixed-step Halpern ($\beta=\tfrac12$, \Cref{prop:fixed-halpern}), and a coda layer
(\Cref{app:looped-shakespeare}).
\Sparrow{} lowers the final validation loss for every $N$ and every truncation window $H$ on all three seeds, by
$0.012$--$0.027$ (\Cref{fig:loop-shakespeare}).  

\section{Perspective}
\label{sec:conclusion}
Taking the Lipschitz certificate approach in order to derive efficient and robust optimization algorithms seems like an unreasonably predictive approach.
Can Lipschitz certificates similarly guide optimizer design for other modules like Mixture-of-Experts (MoE)?
How should we be scaling models, with optimizers that properly utilize parameters of tall matrices?
Are there any model design-choices we discarded in the past because we did not optimize them appropriately?

Regarding recurrent-depth, the grounding of input injection in Halpern iteration makes it possible to ask a range of questions:
Is there a recurrence that converges to a fixed point of
$\opT_\theta$, like slowly vanishing Halpern, which also tolerates operators
that are not strictly nonexpansive, like fixed-step Halpern?
Is it worthwhile letting go of truncated backpropagation to support a parameterization like $\beta_k=1/(k+2)$ which enjoys faster convergence rates and thus shorter depth?
\looseness=-1

\section*{AI use statement}
The desiderata, the connection between input injection and Halpern iteration, its contractive property and the consequences for truncated backpropagation, the observations surrounding robustness, and the central Lipschitz argument for squared ReLU and gating were developed by the author.
Generative AI were used to extend this argument to SwiGLU and state-space models, and an LLM proposed the normalization $\opN_{\tanh}$.
The connection to forward-backward splitting and iDRS, and the practical Sparrow optimizer, were developed by the author, while the low-rank case in \Cref{app:lmo-idrs} was proposed by generative AI.
Generative AI were used extensively for formalizing arguments, writing code and producing plots.
The author takes full responsibility for the final content of this work, including text, claims or artifacts produced with the aid of generative AI.

\section*{Acknowledgments}\label{sec:acknowledgments}
The author thanks the Swiss chômage for supporting the work.

\bibliographystyle{plainnat}
\bibliography{refs}

\clearpage
\appendix
\etocdepthtag.toc{appendix}
\etocsettagdepth{mainmatter}{none}
\etocsettagdepth{appendix}{subsection}
\etocsettocstyle{\section*{Contents of the appendix}}{}
\tableofcontents
\section{Proofs and supplementary results for Halpern iteration}
\label[appendix]{app:halpern-iteration}
\subsection{Preliminary: geometric TBPTT error}
\begin{lemma}[Geometric TBPTT error]
\label{lem:tbptt}
Suppose the recurrence is differentiable along its trajectory, $u$ and
$z^0$ are independent of $\theta$, and
\[
A_k:=\partial_z\opG_{k,\theta,u}(z^k),
\qquad
B_k:=\partial_\theta \opG_{k,\theta,u}(z^k)
\]
satisfy $\|A_k\|\le q<1$ and $\|B_k\|\le B$ for every $k$.  For
$0\le H\le K$, let
\[
S_K^{(H)}
:=
\textstyle\sum_{i=K-H}^{K-1}A_{K-1}\cdots A_{i+1}B_i,
\]
where an empty product is the identity.  Then, with
$S_K:=\partial_\theta z^K$,
\begin{equation}
\|S_K-S_K^{(H)}\|
\le \tfrac{B}{1-q}q^H.
\label{eq:generic-tbptt}
\end{equation}
In particular, a horizon
$H=O(\log(1/\varepsilon))$ gives error at most $\varepsilon$,
independently of $K$.
\end{lemma}

\begin{proof}
The chain rule gives $S_{k+1}=A_kS_k+B_k$ and $S_0=0$.  Iterating,
\[
S_K=\textstyle\sum_{i=0}^{K-1}A_{K-1}\cdots A_{i+1}B_i.
\]
Therefore
\[
\|S_K-S_K^{(H)}\|
\le
B\textstyle\sum_{i=0}^{K-H-1}q^{K-1-i}
=B\textstyle\sum_{j=H}^{K-1}q^j
\le \tfrac{B}{1-q}q^H.
\qedhere
\]
\end{proof}

\subsection{Fixed-step Halpern}
\begin{proof}[Proof of \Cref{prop:fixed-halpern}]
Since $\opT_\theta$ is $L$-Lipschitz,
\[
\|\opH_{\beta,u}(x)-\opH_{\beta,u}(y)\|
=(1-\beta)\|\opT_\theta(x)-\opT_\theta(y)\|
\le q\|x-y\|,
\]
so $\opH_{\beta,u}$ is a contraction with modulus $q$, and (i) follows
from the Banach fixed-point theorem.  At every point of differentiability,
$\|\partial_z\opH_{\beta,u}\|\le q$, so (ii) is \Cref{lem:tbptt} with
this $q$.

For (iii), rearranging the fixed-point equation
$z_\beta=\beta u+(1-\beta)\opT_\theta(z_\beta)$ gives
\begin{equation}
z_\beta-\opT_\theta(z_\beta)=\tfrac{\beta}{1-\beta}(u-z_\beta),
\label{eq:fixed-halpern-residual}
\end{equation}
which is the residual identity.  Since $\beta>0$, it shows that
$z_\beta\in\Fix(\opT_\theta)$ if and only if $z_\beta=u$.  Hence, if
$z_\beta$ is a fixed point, then $u=z_\beta\in\Fix(\opT_\theta)$.
Conversely, if $u\in\Fix(\opT_\theta)$, then $\opH_{\beta,u}(u)=u$, and
uniqueness gives $z_\beta=u$.

Now let $\opT_\theta$ be nonexpansive, so that $\opI-\opT_\theta$ is
monotone~\citep{bauschke2017}, and let $c\in\cC_\theta$.  Since $(\opI-\opT_\theta)c=0$,
monotonicity and \eqref{eq:fixed-halpern-residual} give
\[
0\le\langle(\opI-\opT_\theta)z_\beta,\,z_\beta-c\rangle
=\tfrac{\beta}{1-\beta}\langle u-z_\beta,\,z_\beta-c\rangle,
\]
that is, $\langle u-z_\beta,c-z_\beta\rangle\le0$.  Consequently,
\[
\|u-c\|^2
=\|u-z_\beta\|^2+\|z_\beta-c\|^2
  +2\langle u-z_\beta,z_\beta-c\rangle
\ge\|u-z_\beta\|^2.
\]
Taking the infimum over $c\in\cC_\theta$ yields
$\|u-z_\beta\|\le\operatorname{dist}(u,\cC_\theta)$, and the residual
identity gives the stated residual bound.
\end{proof}

\subsection{Logarithmic-step Halpern}
\begin{proof}[Proof of \Cref{prop:log-halpern}]
The sequence $(\beta_k)$ lies in $(0,1)$, decreases to zero, and
\[
\textstyle\sum_{k=0}^\infty\beta_k=\infty,
\qquad
\textstyle\sum_{k=0}^\infty|\beta_{k+1}-\beta_k|
=\beta_0<\infty.
\]
The divergence follows, for example, from
$\log(k+3)\le k+3$ and comparison with the harmonic series; the second
identity follows by monotonicity and telescoping.  The classical Halpern
convergence theorem~\citep{wittmann1992} therefore proves (i).

For the sensitivity estimates in (ii), write
\[
S_k:=\partial_\theta z^k,
\qquad
A_k:=(1-\beta_k)\partial_z\opT_\theta(z^k),
\qquad
B_k:=(1-\beta_k)\partial_\theta \opT_\theta(z^k).
\]
Then $S_{k+1}=A_kS_k+B_k$, with
\[
\|A_k\|\le1-\beta_k,
\qquad
\|B_k\|\le M.
\]
For every $n\ge1$, iteration of this recursion and monotonicity of
$(\beta_k)$ give
\begin{align}
\|S_n\|
&\le
M\textstyle\sum_{i=0}^{n-1}
\textstyle\prod_{j=i+1}^{n-1}(1-\beta_j)\notag\\
&\le
M\textstyle\sum_{r=0}^{n-1}(1-\beta_{n-1})^r
\le\tfrac{M}{\beta_{n-1}}
=M\log(n+2).
\label{eq:log-halpern-full-sensitivity}
\end{align}
The part discarded by $H$-step TBPTT is exactly
\[
S_K-S_K^{(H)}
=A_{K-1}\cdots A_{K-H}S_{K-H}.
\]
For $H<K$, using $1-t\le e^{-t}$ and the fact that
$\beta_j\ge\beta_{K-1}=1/\log(K+2)$ when $j\le K-1$, we obtain
\begin{align*}
\|S_K-S_K^{(H)}\|
&\le
\|S_{K-H}\|
\textstyle\prod_{j=K-H}^{K-1}(1-\beta_j)\\
&\le
M\log(K-H+2)
\exp\!\left(-\textstyle\sum_{j=K-H}^{K-1}\beta_j\right)\\
&\le
M\log(K+2)
\exp\!\left(-\tfrac{H}{\log(K+2)}\right),
\end{align*}
which proves (ii).
\end{proof}

\begin{remark}[Bounded full sensitivities]
\label{rem:log-halpern-bounded-sensitivity}
If the full sensitivities also satisfy
$\sup_k\|\partial_\theta z^k\|\le S$, replacing
\eqref{eq:log-halpern-full-sensitivity} by $\|S_{K-H}\|\le S$ in the
proof above gives
\[
\|S_K-S_K^{(H)}\|
\le
S\exp\!\left(-\tfrac{H}{\log(K+2)}\right),
\]
so a window $H=O(\log K\,\log(S/\varepsilon))$ suffices.  This improves
on \Cref{prop:log-halpern}(ii) only if $S$ is smaller than the bound
$\|\partial_\theta z^k\|\le M\log(k+2)$ that
\eqref{eq:log-halpern-full-sensitivity} already provides under the
assumptions of the proposition.
\end{remark}

\subsection{Input robustness}
\begin{proof}[Proof of \Cref{thm:halpern-robustness}]
Write $d_k:=\|z^k-\widetilde z^k\|$.  Subtracting the two updates and
using that $\opT_\theta$ is $L$-Lipschitz gives
\begin{equation}
d_{k+1}
\le\beta_k\|u-\widetilde u\|+(1-\beta_k)L\,d_k
\le\beta_k\varepsilon+(1-\beta_k)L\,d_k.
\label{eq:robustness-one-step}
\end{equation}
If $d_k\le C\varepsilon$, then \eqref{eq:robustness-one-step} gives
\[
d_{k+1}\le\big(\beta_k+(1-\beta_k)LC\big)\varepsilon,
\]
which is at most $C\varepsilon$ provided
\begin{equation}
\beta_k\le C\big(1-(1-\beta_k)L\big).
\label{eq:robustness-invariance}
\end{equation}
For $L\le1$ and $C=1$, \eqref{eq:robustness-invariance} reads
$(1-\beta_k)L\le1-\beta_k$, which holds.  For $L>1$, the condition
$\beta_k\ge\beta_{\min}>1-1/L$ makes $1-(1-\beta_k)L>0$, so
\eqref{eq:robustness-invariance} reads $f(\beta_k)\le C$ for
\[
f(\beta):=\tfrac{\beta}{1-(1-\beta)L},
\qquad
f'(\beta)=\tfrac{1-L}{\big(1-(1-\beta)L\big)^2}<0
\qquad\text{on }(1-1/L,1].
\]
Thus $f$ is decreasing with $f(1)=1$, so
$f(\beta_k)\le f(\beta_{\min})=C$.  In both cases $C\ge1$, so
$d_0\le\varepsilon\le C\varepsilon$, and induction proves
\eqref{eq:trajectory-robustness}.

It remains to show that the constant $C$ cannot be improved, that is,
that there is an operator and a schedule for which the separation $d_k$
approaches $C\varepsilon$.  Take a fixed schedule $\beta_k\equiv\beta$,
so that $\beta_{\min}=\beta$, and $\opT_\theta=L\opI$ with $L>1$ and
$(1-\beta)L<1$.  Solving the fixed-point equation
$z=\beta u+(1-\beta)Lz$ gives
\[
z_\beta=\tfrac{\beta}{1-(1-\beta)L}\,u=C u,
\qquad\text{so}\qquad
\|z_\beta-\widetilde z_\beta\|=C\|u-\widetilde u\|,
\]
where $C$ is the constant \eqref{eq:robustness-constant} for this
schedule.  By \Cref{prop:fixed-halpern}(i), $z^k\to z_\beta$ and
$\widetilde z^k\to\widetilde z_\beta$, so $d_k\to C\|u-\widetilde u\|$,
which equals $C\varepsilon$ when $\|u-\widetilde u\|=\varepsilon$.
Hence, for any $C'<C$, the bound $d_k\le C'\varepsilon$ fails for all
sufficiently large $k$, so \eqref{eq:trajectory-robustness} does not hold
with a smaller constant.
\end{proof}

\section{Proofs and supplementary results for the neural parameterization}
\label[appendix]{app:neural-parameterization}

\subsection{Normalization profiles and the MLP block}
The normalizations of \Cref{sec:lipschitz-parameterization} are
instances of the following construction.

\begin{definition}[Normalization profile]
\label{def:normalization-profile}
Let $\tau>0$.  A \emph{normalization profile} is a concave, continuously
differentiable $h:[0,\infty)\to[0,\tau]$ with $h(0)=0$ and $h'(0)\le1$.
It defines the radial normalization and constant
\[
\opN_h(x):=\tfrac{h(\operatorname{rms}(x))}{\operatorname{rms}(x)}\,x,
\quad\opN_h(0):=0,
\qquad
\omega_h:=\sup_{\rho>0}\tfrac{h(\rho)^2}{\rho}.
\]
We call $\opN_h$ a \emph{normalization layer}.
\end{definition}

\begin{lemma}[Properties of normalization layers]
\label{lem:general-profile}
For every normalization profile $h$, $\opN_h$ is continuously
differentiable away from $0$, and for all $x,y\in\mathbb{R}^d$:
\begin{enumerate}[label=(N\arabic*),leftmargin=3em]
    \item\label{item:N1} $\|\opN_h(x)-\opN_h(y)\|\le\|x-y\|$;
    \item\label{item:N2} $\|\opN_h(x)\|\le\sqrt d\,\tau$;
    \item\label{item:N3} $\|\opN_h(x)\|\,\|D\opN_h(x)\|_{\mathrm{op}}\le\sqrt d\,\omega_h$
    for $x\ne0$, with $\omega_h\le\tau$.
\end{enumerate}
\end{lemma}

\begin{proof}
By concavity, $h'$ is nonincreasing, and $h'\ge0$ since otherwise
$h(\rho)\to-\infty$ as $\rho\to\infty$, contradicting $h\ge0$.  Hence
$\rho\,h'(\rho)\le h(\rho)=\int_0^\rho h'(s)\,ds\le\rho\,h'(0)\le\rho$,
that is,
\begin{equation}
0\le h'(\rho)\le\tfrac{h(\rho)}{\rho}\le1
\qquad\text{for every }\rho>0.
\label{eq:profile-slopes}
\end{equation}

Fix $x\ne0$, and write $\hat x:=x/\|x\|$ and $\rho:=\operatorname{rms}(x)$,
so that $\opN_h(x)=\sqrt d\,h(\rho)\hat x$.  Differentiating the radial
factor and the direction separately gives
\[
D\opN_h(x)
=h'(\rho)\,\hat x\hat x^\top+\tfrac{h(\rho)}{\rho}\big(\opI-\hat x\hat x^\top\big),
\]
which is continuous in $x\ne0$.  This symmetric matrix has eigenvalue
$h'(\rho)$ along $\hat x$ and $h(\rho)/\rho$ on its orthogonal complement,
so \eqref{eq:profile-slopes} gives
\begin{equation}
\|D\opN_h(x)\|_{\mathrm{op}}=\tfrac{h(\rho)}{\rho}\le1.
\label{eq:profile-jacobian}
\end{equation}

For \ref{item:N1}, $\opN_h$ is continuous and a nonconstant line segment meets $0$
at most once, so integrating \eqref{eq:profile-jacobian} along the segment
between $x$ and $y$ gives $\|\opN_h(x)-\opN_h(y)\|\le\|x-y\|$.  For \ref{item:N2},
$\|\opN_h(x)\|=\sqrt d\,h(\rho)\le\sqrt d\,\tau$.  For \ref{item:N3},
\eqref{eq:profile-jacobian} gives
\[
\|\opN_h(x)\|\,\|D\opN_h(x)\|_{\mathrm{op}}
=\sqrt d\,\tfrac{h(\rho)^2}{\rho}
\le\sqrt d\,\omega_h,
\]
and $h(\rho)^2/\rho\le h(\rho)\le\tau$ by \eqref{eq:profile-slopes}, so
$\omega_h\le\tau$.
\end{proof}

\begin{figure}[t]
    \centering
    \includegraphics[width=\linewidth]{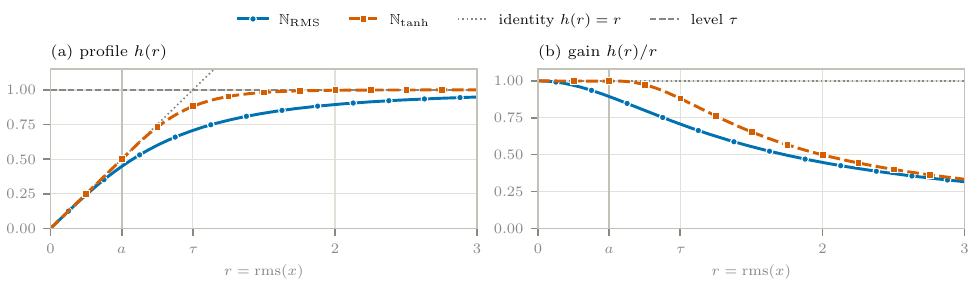}
    \caption{Profiles of the normalization layers $\opN_{\mathrm{RMS}}$ and
    $\opN_{\tanh}$ in \eqref{eq:profiles} with $\tau=1$ and $a=\tfrac12$, as
    functions of $r=\operatorname{rms}(x)$: (a) the profile $h(r)$ and (b) the
    gain $h(r)/r=\|D\opN_h(x)\|_{\mathrm{op}}$ \eqref{eq:profile-jacobian}.
    Both are nonexpansive, since their gains stay at most $1$.  The gain of
    $\opN_{\tanh}$ equals $1$ up to $r=a$ so its
    compositions with other nonexpansive maps can have multiple fixed points.}
    \label{fig:norm-profiles}
\end{figure}

\Cref{fig:norm-profiles} shows the profiles of $\opN_{\mathrm{RMS}}$ and
$\opN_{\tanh}$ in \eqref{eq:profiles}, $h(r)=\tau r/\sqrt{\tau^2+r^2}$ and
$h(r)=r$ for $r\le a$, $a+(\tau-a)\tanh\big(\tfrac{r-a}{\tau-a}\big)$ for $r>a$.

The block theorems of \Cref{sec:lipschitz-parameterization} use
only \ref{item:N1}--\ref{item:N3}, so they hold for every normalization layer, and indeed
for any map with these three properties.

\begin{lemma}[The two normalizations]
\label{lem:profile-properties}
$\opN_{\mathrm{RMS}}$ and $\opN_{\tanh}$ are normalization layers.  They
satisfy \ref{item:N1}--\ref{item:N3} with $\omega=\tau/2$ and $a\le\omega\le\tau$,
respectively, and $\opN_{\mathrm{RMS}}$ strictly decreases every nonzero
distance.
\end{lemma}

\begin{proof}
By \Cref{lem:general-profile}, it suffices to write each normalization as
$\opN_h$, check that $h$ is a normalization profile, and compute
$\omega=\omega_h$.

First, $\opN_{\mathrm{RMS}}=\opN_h$ with
\[
h(\rho)=\tfrac{\tau\rho}{\sqrt{\tau^2+\rho^2}},
\qquad
h'(\rho)=\tfrac{\tau^3}{(\tau^2+\rho^2)^{3/2}}.
\]
Thus $h(0)=0$, $h<\tau$, and $h'$ is continuous and decreases from
$h'(0)=1$, so $h$ is a normalization profile.  Its constant is
\[
\omega=\sup_{\rho>0}\tfrac{\tau^2\rho}{\tau^2+\rho^2}=\tfrac\tau2,
\]
attained at $\rho=\tau$.  Moreover, $h(\rho)/\rho=\tau/\sqrt{\tau^2+\rho^2}<1$
for $\rho>0$, so \eqref{eq:profile-jacobian} is strict at every $x\ne0$.
Integrating along the segment between $x\ne y$, as in the proof of \ref{item:N1},
therefore gives
\[
\|\opN_{\mathrm{RMS}}(x)-\opN_{\mathrm{RMS}}(y)\|<\|x-y\|,
\]
which is the strict decrease used for the normalization layers in
\Cref{sec:lipschitz-parameterization}.

Second, $\opN_{\tanh}=\opN_h$ with
\[
h(\rho)=\min(\rho,a)+(\tau-a)\tanh\!\Big(\tfrac{(\rho-a)_+}{\tau-a}\Big),
\qquad
h'(\rho)=
\begin{cases}
1, & \rho\le a,\\
\operatorname{sech}^2\!\big(\tfrac{\rho-a}{\tau-a}\big), & \rho>a.
\end{cases}
\]
Thus $h(0)=0$, $h<\tau$, and $h'$ is continuous and nonincreasing from
$h'(0)=1$, so $h$ is a normalization profile.  Since $h(\rho)^2/\rho=\rho$
on $(0,a]$, its constant satisfies $\omega\ge a$, and $\omega\le\tau$ by
\Cref{lem:general-profile}.
\end{proof}

\begin{proof}[Proof of \Cref{thm:nonexpansive-neural-block}]
The raw MLP satisfies
\[
\|\opF_{\mathrm{MLP}}(x)-\opF_{\mathrm{MLP}}(\widetilde x)\|
\le
\|W_2\|_{\mathrm{op}}L_\sigma\|W_1\|_{\mathrm{op}}
\|x-\widetilde x\|,
\]
where the biases cancel in the difference.  Since $\opN$ is
$1$-Lipschitz by \ref{item:N1}, composing it before or after
$\opF_{\mathrm{MLP}}$ does not increase this bound, which proves
\eqref{eq:mlp-lipschitz}.
\end{proof}

\subsection{Linearly growing Jacobians, SwiGLU, and squared ReLU}
\label[appendix]{app:linear-jacobian}
\begin{lemma}[Input normalization for linearly growing Jacobians]
\label{lem:input-squash-linear-jacobian}
Let $\opG:\mathbb{R}^d\to\mathbb{R}^d$ be continuously differentiable,
and suppose that, for some $L_1,L_0\ge0$,
\begin{equation}
\|D\opG(z)\|_{\mathrm{op}}\le L_1\|z\|+L_0
\qquad\text{for every }z\in\mathbb{R}^d.
\label{eq:linear-jacobian-growth}
\end{equation}
Then, for every normalization layer $\opN$
(\Cref{def:normalization-profile}),
\begin{equation}
\operatorname{Lip}(\opG\circ\opN)
\le L_1\sqrt d\,\omega+L_0.
\label{eq:linear-jacobian-squash-bound}
\end{equation}
For $\opN_{\mathrm{RMS}}$, the bound is $L_1\sqrt d\,\tau/2+L_0$.
\end{lemma}

\begin{proof}
At $x\ne0$, the chain rule, \ref{item:N3} and $\|D\opN(x)\|_{\mathrm{op}}\le1$,
which follows from \ref{item:N1}, give
\[
\|D(\opG\circ\opN)(x)\|_{\mathrm{op}}
\le
\big(L_1\|\opN(x)\|+L_0\big)\|D\opN(x)\|_{\mathrm{op}}
\le
L_1\sqrt d\,\omega+L_0.
\]
Integrating the Jacobian along line segments, each of which meets $0$ at
most once, proves \eqref{eq:linear-jacobian-squash-bound}.
\end{proof}
We are now ready to make the main argument that makes us arrive at a joint spectral and row constraint for matrices followed by ReLU$^2$ and gated activation functions such as in SwiGLU.

\begin{theorem}[Lipschitz squared-ReLU block]
\label{thm:nonexpansive-relu2}
Let $W_1:\mathbb{R}^d\to\mathbb{R}^m$ and
$W_2:\mathbb{R}^m\to\mathbb{R}^d$ be linear maps, with
$b_1\in\mathbb{R}^m$ and $b_2\in\mathbb{R}^d$, let $\opN$ be a
normalization layer (\Cref{def:normalization-profile}), and define
\begin{equation}
\opM_{\mathrm{ReLU}^2}(x)
:=
W_2\operatorname{ReLU}\!\left(W_1\opN(x)+b_1\right)^{\odot2}
+b_2.
\label{eq:relu2-map}
\end{equation}
Then
\begin{equation}
\operatorname{Lip}(\opM_{\mathrm{ReLU}^2})
\le
2\|W_2\|_{\mathrm{op}}\|W_1\|_{\mathrm{op}}
\left[
\sqrt d\,\omega\|W_1\|_{2\to\infty}
+\|\operatorname{ReLU}(b_1)\|_\infty
\right].
\label{eq:relu2-lipschitz}
\end{equation}
\end{theorem}

\begin{proof}
Define
\[
\opH(z)
:=
W_2\operatorname{ReLU}(W_1z+b_1)^{\odot2}+b_2.
\]
Since $t\mapsto\operatorname{ReLU}(t)^2$ has derivative
$2\operatorname{ReLU}(t)$,
\[
 D\opH(z)
 =2W_2\operatorname{diag}\!\left(\operatorname{ReLU}(W_1z+b_1)\right)W_1.
\]
The operator norm of a diagonal matrix is the vector $\ell_\infty$ norm over the entries.  Moreover, Cauchy--Schwarz applied to each row gives
\[
\|W_1z\|_\infty
\le\max_i\|(W_1)_{i,:}\|_2\,\|z\|
=\|W_1\|_{2\to\infty}\|z\|.
\]
Using $\operatorname{ReLU}(u+v)\le |u|+\operatorname{ReLU}(v)$
coordinatewise, we therefore obtain
\[
\begin{aligned}
\|D\opH(z)\|_{\mathrm{op}}
&\le
2\|W_2\|_{\mathrm{op}}\|W_1\|_{\mathrm{op}}
\|\operatorname{ReLU}(W_1z+b_1)\|_\infty\\
&\le
2\|W_2\|_{\mathrm{op}}\|W_1\|_{\mathrm{op}}
\left(
\|W_1\|_{2\to\infty}\|z\|
+\|\operatorname{ReLU}(b_1)\|_\infty
\right).
\end{aligned}
\]
Thus its Jacobian has the linear-growth structure of \Cref{lem:input-squash-linear-jacobian}.  Applying \Cref{lem:input-squash-linear-jacobian} with
\[
L_1=2\|W_2\|_{\mathrm{op}}\|W_1\|_{\mathrm{op}}
\|W_1\|_{2\to\infty},
\qquad
L_0=2\|W_2\|_{\mathrm{op}}\|W_1\|_{\mathrm{op}}
\|\operatorname{ReLU}(b_1)\|_\infty
\]
proves \eqref{eq:relu2-lipschitz}.
\end{proof}

\begin{proof}[Proof of \Cref{thm:nonexpansive-swiglu}]
Let $z=\opN(x)$ and
$\opG(z):=W_o[\operatorname{SiLU}(W_gz)\odot W_vz]$.  The product rule
gives
\[
D\opG(z)
=W_o\left[
\operatorname{Diag}\!\left(\operatorname{SiLU}'(W_gz)\odot W_vz\right)W_g
+\operatorname{Diag}\!\left(\operatorname{SiLU}(W_gz)\right)W_v
\right].
\]
Since $|\operatorname{SiLU}(t)|\le|t|$ and
$|\operatorname{SiLU}'(t)|\le L_{\mathrm{SiLU}}$, the row-norm bounds
$\|W_gz\|_\infty\le\|W_g\|_{2\to\infty}\|z\|$ and
$\|W_vz\|_\infty\le\|W_v\|_{2\to\infty}\|z\|$ imply
\begin{equation}
\|D\opG(z)\|_{\mathrm{op}}
\le C\|z\|,
\quad
C:=\|W_o\|_{\mathrm{op}}
\left[
L_{\mathrm{SiLU}}\|W_v\|_{2\to\infty}\|W_g\|_{\mathrm{op}}
+\|W_g\|_{2\to\infty}\|W_v\|_{\mathrm{op}}
\right].
\label{eq:swiglu-derivative-bound}
\end{equation}
Thus \eqref{eq:linear-jacobian-growth} in \Cref{lem:input-squash-linear-jacobian} holds with $L_1=C$ and $L_0=0$.
\Cref{lem:input-squash-linear-jacobian} gives
\eqref{eq:swiglu-lipschitz}.
\end{proof}

\begin{remark}[Alternative SwiGLU certificates]
\label{rem:swiglu-alternatives}
Applied to a direction $h$, the derivative in the proof above is
\[
D\opG(z)h=W_o\big[(W_vz)\odot\big(\operatorname{SiLU}'(W_gz)\odot W_gh\big)
+\operatorname{SiLU}(W_gz)\odot(W_vh)\big].
\]
Each product $a\odot b$ can be bounded by $\|a\|_\infty\|b\|$ or by
$\|a\|\|b\|_\infty$, and the $\ell_\infty$ norm of a projection is controlled
by its row norm.  With $r_\cdot:=\|W_\cdot\|_{2\to\infty}$ and
$s_\cdot:=\|W_\cdot\|_{\mathrm{op}}$, the four choices give the constants
$C/\|W_o\|_{\mathrm{op}}$ below, where (i) is the one used in
\Cref{thm:nonexpansive-swiglu}.
\begin{center}
\small
\begin{tabular}{@{}lllll@{}}
\toprule
& $\ell_\infty$ in first term & $\ell_\infty$ in second term & $C/\|W_o\|_{\mathrm{op}}$ & Norms used \\
\midrule
(i) & $W_vz$ & $\operatorname{SiLU}(W_gz)$ & $L_{\mathrm{SiLU}}r_vs_g+r_gs_v$ & rows and spectra of both \\
(ii) & $W_gh$ & $\operatorname{SiLU}(W_gz)$ & $(1+L_{\mathrm{SiLU}})r_gs_v$ & row of $W_g$, spectrum of $W_v$ \\
(iii) & $W_vz$ & $W_vh$ & $(1+L_{\mathrm{SiLU}})r_vs_g$ & row of $W_v$, spectrum of $W_g$ \\
(iv) & $W_gh$ & $W_vh$ & $L_{\mathrm{SiLU}}r_gs_v+r_vs_g$ & rows and spectra of both \\
\bottomrule
\end{tabular}
\end{center}
The four coincide when $r_v/s_v=r_g/s_g$, in particular for matrices
saturated in both norms, where each equals $(1+L_{\mathrm{SiLU}})rs$.  This
is the case used in \Cref{rem:swiglu-budget} to choose the predetermined radii
of \Sparrow{}.
\end{remark}

\subsection{The Sinkhorn-attention block}
\label[appendix]{app:sinkhorn-attention}
We use Sinkhorn based attention, following Sinkformers~\citep{sander2022}, since its doubly stochastic attention matrices are nonexpansive.
Equip $\mathbb{R}^{n\times d}$ with the
Frobenius norm.  Write
$
\mathcal B_n
:=
\{P\in\mathbb{R}^{n\times n}_+:
P\mathbf 1=\mathbf 1,\ P^\top\mathbf 1=\mathbf 1\}
$
for the Birkhoff polytope, and define the exact entropy-regularized
Sinkhorn map by
\looseness=-1
\begin{equation}
\operatorname{Sink}(C)
:=
\operatorname*{arg\,max}_{P\in\mathcal B_n}
\left\{\langle C,P\rangle_F-\textstyle\sum_{i,j}P_{ij}\log P_{ij}\right\}.
\label{eq:sinkhorn-map}
\end{equation}

\begin{theorem}[Lipschitz Sinkhorn-attention block]
\label{thm:nonexpansive-attention}
Let $W_Q,W_K:\mathbb{R}^d\to\mathbb{R}^{d_k}$,
$W_V:\mathbb{R}^d\to\mathbb{R}^{d_v}$, and
$W_O:\mathbb{R}^{d_v}\to\mathbb{R}^d$ be linear maps.  Fix a
normalization layer $\opN$ (\Cref{def:normalization-profile}), and positive spectral upper
bounds $\kappa_Q\ge\|W_Q\|_{\mathrm{op}}$ and
$\kappa_K\ge\|W_K\|_{\mathrm{op}}$.  Define
\begin{equation}
\opZ(X):=\opN(X),
\qquad
\opQ(X):=\opZ(X)W_Q,
\qquad
\opK(X):=\opZ(X)W_K,
\qquad
\opV(X):=\opZ(X)W_V.
\label{eq:shared-input-qkv}
\end{equation}
For an
inverse-temperature parameter $B>0$, set
\begin{equation}
\opC(X):=\tfrac{B\,\opQ(X)\opK(X)^\top}
{d\tau^2\kappa_Q\kappa_K},
\qquad
\opW(X):=\operatorname{Sink}(\opC(X)).
\label{eq:sinkhorn-score-map}
\end{equation}
Then $|[\opC(X)]_{ij}|\le B$.  Define the attention block by
\begin{equation}
\opA(X)
:=\opW(X)\opV(X)W_O.
\label{eq:attention-map}
\end{equation}
Then, in the Frobenius norm,
\begin{equation}
\operatorname{Lip}(\opA)
\le
\left[
1+Be^{4B}
\left(
\tfrac{\|W_Q\|_{\mathrm{op}}}{\kappa_Q}
+\tfrac{\|W_K\|_{\mathrm{op}}}{\kappa_K}
\right)
\right]
\|W_V\|_{\mathrm{op}}\|W_O\|_{\mathrm{op}}
\le(1+2Be^{4B})\|W_V\|_{\mathrm{op}}\|W_O\|_{\mathrm{op}},
\label{eq:sinkhorn-attention-lipschitz}
\end{equation}
independently of the sequence length $n$.
\end{theorem}

\begin{proof}[Proof of \Cref{thm:nonexpansive-attention}]
We first record the sensitivity of exact Sinkhorn normalization on the
bounded score range.  Let $P=\operatorname{Sink}(C)$ with $|C_{ij}|\le B$.
Since $P$ is a diagonal scaling of
$\exp(C)$~\citep{sinkhorn1967,peyre2019computational}, write
$P_{ij}=a_i e^{C_{ij}}b_j$.  The column constraints give
\[
\tfrac{b_j}{b_k}
=
\tfrac{\textstyle\sum_i a_i e^{C_{ik}}}
{\textstyle\sum_i a_i e^{C_{ij}}}
\le e^{2B},
\]
hence $P_{ij}/P_{ik}\le e^{4B}$.  Because each row sums to one,
\begin{equation}
0<P_{ij}\le
\rho_n(B)
:=
\tfrac{e^{4B}}{n-1+e^{4B}}.
\label{eq:sinkhorn-entry-bound}
\end{equation}
Now let $\widetilde P=\operatorname{Sink}(\widetilde C)$ with
$|\widetilde C_{ij}|\le B$, and let
$\Phi(P):=\textstyle\sum_{i,j}P_{ij}(\log P_{ij}-1)$, which differs from
the regularizer in \eqref{eq:sinkhorn-map} by the constant $n$ on
$\mathcal B_n$.  The optimality conditions for \eqref{eq:sinkhorn-map} at
$P$ and $\widetilde P$, tested with $\widetilde P$ and $P$ respectively,
give
\[
\langle C-\nabla\Phi(P),\widetilde P-P\rangle\le0,
\qquad
\langle \widetilde C-\nabla\Phi(\widetilde P),P-\widetilde P\rangle\le0.
\]
By \eqref{eq:sinkhorn-entry-bound}, the Hessian
$\nabla^2\Phi=\operatorname{diag}(1/P_{ij})$ is at least
$\rho_n(B)^{-1}\opI$ on the segment between $P$ and $\widetilde P$.
Adding the two conditions and using Cauchy--Schwarz therefore gives
\[
\rho_n(B)^{-1}\|P-\widetilde P\|_F^2
\le\langle\nabla\Phi(P)-\nabla\Phi(\widetilde P),P-\widetilde P\rangle
\le\langle C-\widetilde C,P-\widetilde P\rangle
\le\|C-\widetilde C\|_F\|P-\widetilde P\|_F,
\]
that is,
\begin{equation}
\|\operatorname{Sink}(C)-\operatorname{Sink}(\widetilde C)\|_F
\le
\rho_n(B)\|C-\widetilde C\|_F.
\label{eq:sinkhorn-lipschitz}
\end{equation}

Write $\Delta:=\|\opZ(X)-\opZ(Y)\|_F$.  By \ref{item:N1} and \ref{item:N2} for the shared
input normalization, applied rowwise, $\Delta\le\|X-Y\|_F$ and
\begin{equation}
\begin{aligned}
&&\|z_i(X)\|&\le\sqrt d\,\tau,\\
\|\opQ(X)-\opQ(Y)\|_F&\le\|W_Q\|_{\mathrm{op}}\Delta,
&\|q_i(X)\|&\le\sqrt d\,\tau\kappa_Q,\\
\|\opK(X)-\opK(Y)\|_F&\le\|W_K\|_{\mathrm{op}}\Delta,
&\|k_i(X)\|&\le\sqrt d\,\tau\kappa_K,\\
\|\opV(X)-\opV(Y)\|_F&\le\|W_V\|_{\mathrm{op}}\Delta,
&\|v_i(X)\|&\le\sqrt d\,\tau\|W_V\|_{\mathrm{op}}.
\end{aligned}
\label{eq:shared-input-qkv-bounds}
\end{equation}
The row bounds give $\|\opQ(X)\|_{\mathrm{op}}\le\|\opQ(X)\|_F\le\sqrt{nd}\,\tau\kappa_Q$,
and likewise for $\opK$ and $\opV$.  Splitting
$\opQ(X)\opK(X)^\top-\opQ(Y)\opK(Y)^\top$ into a difference in $\opQ$ and
a difference in $\opK$ and dividing by $d\tau^2\kappa_Q\kappa_K$ then gives
\begin{equation}
\|\opC(X)-\opC(Y)\|_F
\le
\tfrac{B\sqrt n}{\sqrt d\,\tau}
\left(
\tfrac{\|W_Q\|_{\mathrm{op}}}{\kappa_Q}
+\tfrac{\|W_K\|_{\mathrm{op}}}{\kappa_K}
\right)
\Delta.
\label{eq:sinkhorn-score-lipschitz}
\end{equation}

For the attention output, split
\[
\opW(X)\opV(X)-\opW(Y)\opV(Y)
=\opW(X)\big(\opV(X)-\opV(Y)\big)+\big(\opW(X)-\opW(Y)\big)\opV(Y).
\]
Since $\opW(X)$ is doubly stochastic, it is a convex combination of
permutation matrices by the Birkhoff--von Neumann theorem, so
$\|\opW(X)\|_{\mathrm{op}}=1$ and the first term is at most
$\|W_V\|_{\mathrm{op}}\Delta$ in the Frobenius norm.  By
\eqref{eq:sinkhorn-lipschitz}, \eqref{eq:sinkhorn-score-lipschitz} and
$\|\opV(Y)\|_{\mathrm{op}}\le\sqrt{nd}\,\tau\|W_V\|_{\mathrm{op}}$, the
second is at most
$Bn\rho_n(B)\big(\tfrac{\|W_Q\|_{\mathrm{op}}}{\kappa_Q}
+\tfrac{\|W_K\|_{\mathrm{op}}}{\kappa_K}\big)\|W_V\|_{\mathrm{op}}\Delta$.
Finally, $n\rho_n(B)=\tfrac{ne^{4B}}{n-1+e^{4B}}\le e^{4B}$, so applying
$W_O$ gives
\begin{equation}
\|\opA(X)-\opA(Y)\|_F
\le
\left[
1+Be^{4B}
\left(
\tfrac{\|W_Q\|_{\mathrm{op}}}{\kappa_Q}
+\tfrac{\|W_K\|_{\mathrm{op}}}{\kappa_K}
\right)
\right]
\|W_V\|_{\mathrm{op}}\|W_O\|_{\mathrm{op}}
\|\opN(X)-\opN(Y)\|_F.
\label{eq:sinkhorn-attention-normalized}
\end{equation}
Since $\|\opN(X)-\opN(Y)\|_F=\Delta\le\|X-Y\|_F$,
this proves the first bound in \eqref{eq:sinkhorn-attention-lipschitz};
the bounds $\|W_Q\|_{\mathrm{op}}\le\kappa_Q$ and
$\|W_K\|_{\mathrm{op}}\le\kappa_K$ give the second.

\end{proof}

\subsection{Gated attention and gated state-space models}
\label[appendix]{app:gated}
Gated attention and gated SSMs both multiply a bounded signal elementwise
by a sigmoid gate computed from the normalized input.  Similarly to the
ReLU$^2$ based MLP and the SwiGLU block (\Cref{thm:nonexpansive-relu2,thm:nonexpansive-swiglu}), the gate acts as a diagonal
matrix, whose operator norm is its largest entry, so its sensitivity ends up being
controlled by the row norms of the gate projection.

\begin{lemma}[Sigmoid gates]
\label{lem:sigmoid-gate}
Let $\alpha(z):=\sigma(Wz+b)$ with $\sigma(t):=(1+e^{-t})^{-1}$.  Then
\begin{equation}
\|\operatorname{Diag}(\alpha(z)-\alpha(\widetilde z))\|_{\mathrm{op}}
=\|\alpha(z)-\alpha(\widetilde z)\|_\infty
\le\tfrac14\|W\|_{2\to\infty}\|z-\widetilde z\|,
\label{eq:sigmoid-gate-row-bound}
\end{equation}
and the bound is unchanged if each gate is repeated across several
channels.
\end{lemma}

\begin{proof}
Since $|\sigma'|\le1/4$, the $i$th gate changes by at most
$\tfrac14|\langle W_{i,:},z-\widetilde z\rangle|
\le\tfrac14\|W_{i,:}\|\|z-\widetilde z\|$.
Repeating entries does not change the maximum.
\end{proof}

Both certificates below split a gated product as
\begin{equation}
D(x)u(x)-D(y)u(y)=D(x)\big(u(x)-u(y)\big)+\big(D(x)-D(y)\big)u(y),
\label{eq:gated-product-split}
\end{equation}
where $D$ is the diagonal gate matrix.  The first term is bounded by the
gate magnitude, and the second by \Cref{lem:sigmoid-gate} times a bound $M$
on $\|u\|$, which comes from the attention heads in gated attention and from
the stability of the recurrence in a gated SSM.
Both blocks are also continuous functions of the normalized input for each
fixed sequence length (for attention, with the Sinkhorn heads of
\Cref{thm:nonexpansive-attention}), so
\Cref{thm:bounded-normalization-fixed-point} guarantees a fixed point
regardless of the Lipschitz bound.

\subsubsection{Gated attention}
\label[appendix]{app:gated-attention}

\begin{theorem}[Lipschitz headwise-gated attention]
\label{thm:nonexpansive-gated-attention}
Let $\opN$ be a normalization layer (\Cref{def:normalization-profile})
applied to each row, and let
$\opU(X)=[\opU_1(X)\mid\cdots\mid\opU_H(X)]\in\mathbb R^{n\times Hd_v}$
be concatenated attention-head outputs with
$\operatorname{Lip}(\opU)\le L_U$ in the Frobenius norm and
$\sup_{X,i}\|\opU(X)_{i,:}\|\le M$.
For a gate projection $W_a\in\mathbb R^{H\times d}$, $b\in\mathbb R^H$, $c>0$, and
an output projection $W_O\in\mathbb R^{Hd_v\times d}$, define
\begin{equation}
g(X):=c\,\sigma\!\left(\opN(X)W_a^\top+\mathbf1b^\top\right),
\qquad
\opA_g(X):=
[g_{:,1}(X)\odot\opU_1(X)\mid\cdots\mid
g_{:,H}(X)\odot\opU_H(X)]W_O,
\label{eq:gated-attention-map}
\end{equation}
where each gate is broadcast across its head's channels.  Then
\begin{equation}
\operatorname{Lip}(\opA_g)
\le c\|W_O\|_{\mathrm{op}}
\left(L_U+\tfrac M4\|W_a\|_{2\to\infty}\right).
\label{eq:gated-attention-lipschitz}
\end{equation}
\end{theorem}

\begin{proof}
At token $i$, apply \eqref{eq:gated-product-split} to the concatenated head
output $u_i$ and the diagonal matrix $D_i$ of its broadcast gates.  The
first term has norm at most $c\|u_i(X)-u_i(Y)\|$, and by
\Cref{lem:sigmoid-gate} and \ref{item:N1} the second has norm at most
$\tfrac c4M\|W_a\|_{2\to\infty}\|x_i-y_i\|$.  Taking the Euclidean norm across tokens, using the
triangle inequality, and applying $W_O$ proves the bound.
\end{proof}

For the Sinkhorn heads of \Cref{thm:nonexpansive-attention},
$L_U\le(1+2Be^{4B})S_V$ and $M\le\sqrt d\,\tau S_V$ with
$S_V:=(\textstyle\sum_h\|W_{V,h}\|_{\mathrm{op}}^2)^{1/2}$, so the bound
is independent of the sequence length.

\subsubsection{Gated state-space models}
\label[appendix]{app:gated-ssm}

In a gated SSM the gate multiplies the previous state, so the bound $M$ on
the gated signal comes from the stability of the recurrence.

\begin{theorem}[Lipschitz gated state-space models]
\label{thm:nonexpansive-gated-ssm}
Let $\opN$ be a normalization layer (\Cref{def:normalization-profile})
on $\mathbb R^d$,
$W_a,W_V\in\mathbb R^{m\times d}$, $b_a\in\mathbb R^m$, and
$W_O\in\mathbb R^{d\times m}$.  Define
$\opS_g(X)=(y^1,\ldots,y^n)$ by
\begin{equation}
z^k:=\opN(x^k),
\qquad
\alpha(z):=\sigma(W_az+b_a),
\qquad
h^k:=\alpha(z^k)\odot h^{k-1}+W_Vz^k,
\qquad
y^k:=W_Oh^k,
\label{eq:gated-ssm-map}
\end{equation}
with $h^0=0$.  Set
\begin{equation}
\rho:=\max_i\sigma\!\left((b_a)_i+\sqrt d\,\tau\|(W_a)_{i,:}\|\right)<1.
\label{eq:gated-ssm-constants}
\end{equation}
Then, in the Frobenius norm,
\begin{equation}
\operatorname{Lip}(\opS_g)
\le
\tfrac{\|W_O\|_{\mathrm{op}}\|W_V\|_{\mathrm{op}}}{1-\rho}
\left(1+\tfrac{\sqrt d\,\tau\|W_a\|_{2\to\infty}}{4(1-\rho)}\right),
\label{eq:gated-ssm-lipschitz}
\end{equation}
independently of the sequence length $n$.
\end{theorem}

\begin{proof}
By \ref{item:N2}, $\|z^k\|\le\sqrt d\,\tau$, so $\|\alpha(z^k)\|_\infty\le\rho$ and
\[
\|h^k\|\le\rho\|h^{k-1}\|+\sqrt d\,\tau\|W_V\|_{\mathrm{op}}.
\]
Since $h^0=0$, induction gives
$\|h^k\|\le M:=\sqrt d\,\tau\|W_V\|_{\mathrm{op}}/(1-\rho)$.

For two inputs, let $e^k:=\|h^k-\widetilde h^k\|$ and
$\delta^k:=\|x^k-\widetilde x^k\|$.  Applying
\eqref{eq:gated-product-split} to $\alpha(z^k)\odot h^{k-1}$,
\[
h^k-\widetilde h^k
=\alpha(z^k)\odot\big(h^{k-1}-\widetilde h^{k-1}\big)
+\big(\alpha(z^k)-\alpha(\widetilde z^k)\big)\odot\widetilde h^{k-1}
+W_V\big(z^k-\widetilde z^k\big).
\]
The three terms are bounded by $\rho e^{k-1}$, by $\tfrac M4\|W_a\|_{2\to\infty}\delta^k$ using
\Cref{lem:sigmoid-gate} and \ref{item:N1}, and by $\|W_V\|_{\mathrm{op}}\delta^k$
using \ref{item:N1}.  Hence
\begin{equation}
e^k\le\rho e^{k-1}+K\delta^k,
\qquad K:=\|W_V\|_{\mathrm{op}}+\tfrac M4\|W_a\|_{2\to\infty}.
\label{eq:gated-ssm-perturbation}
\end{equation}

Since $e^0=0$, iterating \eqref{eq:gated-ssm-perturbation} gives
$e^k\le K\textstyle\sum_{j\le k}\rho^{k-j}\delta^j$, the convolution of
$(K\delta^k)$ with $(1,\rho,\rho^2,\ldots)$, whose $\ell_1$ norm is
$1/(1-\rho)$.  Young's convolution inequality then yields
\[
\|\opS_g(X)-\opS_g(\widetilde X)\|_F
\le\|W_O\|_{\mathrm{op}}\|(e^k)\|_2
\le\tfrac{\|W_O\|_{\mathrm{op}}K}{1-\rho}
\|X-\widetilde X\|_F,
\]
and substituting $M$ and $K$ proves the claim.
\end{proof}

\paragraph{Gate constraint and stability margin}
To certify a Lipschitz constant $L$, choose a row-norm radius
$R\ge0$ and a gate cap $\bar\rho\in(0,1)$ such that
\begin{equation}
\begin{gathered}
\|W_a\|_{2\to\infty}\le R,
\qquad
\max_i(b_a)_i+\sqrt d\,\tau R
\le\log\tfrac{\bar\rho}{1-\bar\rho},\\
\tfrac{\|W_O\|_{\mathrm{op}}\|W_V\|_{\mathrm{op}}}{1-\bar\rho}
\left(1+\tfrac{\sqrt d\,\tau R}{4(1-\bar\rho)}\right)\le L.
\end{gathered}
\label{eq:gated-ssm-budget}
\end{equation}
The bias condition ensures $\rho\le\bar\rho$, so the certificate holds
uniformly over these constraints.  As in gated attention, $W_a$ needs only a
row-norm constraint, which here controls both the gate sensitivity and the
stability margin.

\paragraph{Input-dependent writes and readouts}
The same proof applies to $h^k=\alpha(z^k)\odot h^{k-1}+\opB(z^k)$ and
$y^k=\opC(z^k)h^k$, provided $\opB$ and $\opC$ are bounded and Lipschitz on
the normalized-input ball.  If $\|\opB\|\le M_B$, the state bound becomes
$M=M_B/(1-\rho)$, and the constant picks up the norm and Lipschitz bounds of
$\opB$ and $\opC$.  This is the structure of selective state-space models such
as Mamba~\citep{gu2023mamba}, whose write and readout depend on the input,
and of matrix memories
$S^k=\operatorname{Diag}(\alpha(z^k))S^{k-1}+k(z^k)v(z^k)^\top$ read out as
$W_O(S^k)^\top q(z^k)$, as in gated linear attention~\citep{yang2024gla} and,
with one gate per head, Mamba-2~\citep{dao2024mamba2}. Vectorizing $S^k$
preserves its norm, and repeating a gate preserves $\|W_a\|_{2\to\infty}$.  These models use
other gate nonlinearities than the sigmoid, which replaces the constant $1/4$
in \Cref{lem:sigmoid-gate} by a bound on the derivative of the gate, and
$\rho$ by the largest gate value on the normalized-input ball.
\looseness=-1

\subsection{Fixed-point existence and uniqueness}
\label[appendix]{app:fixed-point-existence}
Normalization layers provide a way of ensuring existence, as the following theorem makes precise.
\looseness=-1
\begin{theorem}[Fixed points from normalization layer]
\label{thm:bounded-normalization-fixed-point}
Let $\opF:\mathcal H\to\mathcal E$ and $\opG:\mathcal E\to\mathcal H$
be continuous maps between finite-dimensional Hilbert spaces, and let
$\opN:\mathcal E\to\mathcal E$ be a normalization layer
(\Cref{def:normalization-profile}).
Its bounded-range property \ref{item:N2} ensures that
$\opT:=\opG\circ\opN\circ\opF$ has a fixed point.
\end{theorem}

\Cref{thm:bounded-normalization-fixed-point} covers the
preceding blocks through the decompositions
\[
\begin{aligned}
\text{MLP:}\quad
&\opM_{\mathrm{in}}
=\opF_{\mathrm{MLP}}\circ\opN\circ\opI,
\quad
\opM_{\mathrm{out}}
=\opI\circ\opN\circ\opF_{\mathrm{MLP}},\\[1mm]
\text{SwiGLU/$\operatorname{ReLU}^2$:}\quad
&\opM_{\mathrm{SwiGLU}}
=\opG_{\mathrm{SwiGLU}}\circ\opN\circ\opI,\\
&\opM_{\mathrm{ReLU}^2}
=\opG_{\mathrm{ReLU}^2}\circ\opN\circ\opI,\\[1mm]
\text{attention:}\quad
&\opA
=\opG_{\mathrm{attn}}\circ\opN\circ\opI.
\end{aligned}
\]
Here the $\opG$ maps denote the corresponding computations after the
normalized input in \eqref{eq:swiglu-map}, \eqref{eq:relu2-map}, and
\eqref{eq:shared-input-qkv}.  All these module families therefore have
a fixed point, regardless of their
Lipschitz constants.

\begin{proof}[Proof of \Cref{thm:bounded-normalization-fixed-point}]
By \ref{item:N2}, choose $\rho<\infty$ with $\|\opN(z)\|\le\rho$ for all $z\in\mathcal E$.
The closed ball $\{z\in\mathcal E:\|z\|\le\rho\}$ is compact, so
continuity of $\opG$ gives
\[
R
:=
\max_{\|z\|\le\rho}\|\opG(z)\|
<\infty.
\]
It follows that $\|\opT(x)\|\le R$ for every $x\in\mathcal H$.
Therefore $\opT$ maps the closed ball of radius $R$ in $\mathcal H$
into itself, and Brouwer's fixed-point theorem gives
$\Fix(\opT)\ne\varnothing$.
\end{proof}

\subsection{Combining attention and MLP blocks}
\begin{proof}[Proof of \Cref{prop:nonexpansive-parallel-block}]
For every $X,Y\in\mathcal H$, the triangle inequality gives
\[
\|\opF_A(X)-\opF_A(Y)\|
\le(1-\alpha_A)\|X-Y\|+\alpha_A\|\opA_\theta(X)-\opA_\theta(Y)\|
\le(1-\alpha_A+\alpha_AL_A)\|X-Y\|,
\]
and likewise for $\opF_M$.  Lipschitz constants multiply under
composition, which gives the bound for $\opT_{\mathrm{ser}}$.  For the
parallel map,
\begin{align*}
\|\opT_{\mathrm{par}}(X)-\opT_{\mathrm{par}}(Y)\|
&\le(1-\alpha)\|\opA_\theta(X)-\opA_\theta(Y)\|
 +\alpha\|\opM_\theta(X)-\opM_\theta(Y)\|\\
&\le\big((1-\alpha)L_A+\alpha L_M\big)\|X-Y\|.
\end{align*}
With $L_A,L_M\le1$, both bounds are at most $1$.
\end{proof}

\section{Computing and using the oracles}
\label[appendix]{app:oracles}
\subsection{Using the LMO inside an optimizer}
\label[appendix]{app:lmo-optimizer}

The LMO only provides the direction of the update.  In practice we combine it
with a Nesterov-type momentum estimator, weight decay and an optional weight
normalization $\mathsf N$:
\begin{equation}
\begin{aligned}
m^{k+1} &= \mu_k m^k+(1-\mu_k)\nabla f(x^k,\xi_k)\\
\bar m^{k+1} &= \bar\mu_k m^{k+1}+(1-\bar\mu_k)\nabla f(x^k,\xi_k)\\
x^{k+1} &= \mathsf{N}\big((1-\lambda_k\gamma_k)x^k-\gamma_k\lmo_{\mathcal K}(\bar m^{k+1})\big)
\end{aligned}
\label{eq:lmo-optimizer}
\end{equation}
Here $\nabla f(x^k,\xi_k)$ is a stochastic gradient, $\gamma_k>0$ is the step
size and $\lambda_k\geq 0$ the weight decay, and $\bar\mu_k=1$ gives standard Heavy ball
momentum.
The map $\mathsf N$ is either the identity or the rescaling to the initial
Frobenius norm,
\begin{equation}
\mathsf N(X):=\|x^0\|_F\,\tfrac{X}{\|X\|_F},
\label{eq:frobenius-normalization}
\end{equation}
as in Hyperball~\citep{wen2026hyperball}, but without normalizing the update
direction.
For a spectral ball $\mathcal K$ this is Muon with decoupled weight decay as studied in \citet{pethick2025scion,liu2025muon},
and for $\mathcal K=\mathcal K_{r,s}$, with the LMO approximated by
\Cref{alg:sparrow}, it is \Sparrow{}.

\subsection{\Sparrow{} with a row-clipping threshold}
\label[appendix]{app:sparrow-c}

Recall from \Cref{sec:rowspectral-solver} that one step of forward-backward
splitting \eqref{eq:fb} on the dual \eqref{eq:lmo-dual}, started from
$A=\mathsf{RowShrink}_c(G)$, gives
$(\mathsf{RowClip}_r\circ\mathsf{Spectral}_s\circ\mathsf{RowClip}_c)(G)$,
and that \Sparrow{} is obtained for $c\to0$ and $U_1$ for $c\to\infty$.  Here we
consider arbitrary $c$.  

\Cref{alg:sparrow-c} replaces $\mathsf{Row}_r$ on the SwiGLU
gate and value matrices in \Cref{alg:sparrow} by $\mathsf{RowClip}_c$.
The threshold $c$ is in the units of $G$, whose scale changes with the
gradient size.  
We therefore set it
relative to the root-mean-square row norm of $G$,
\[
\bar g:=\Big(\tfrac1m\textstyle\sum_{i=1}^m\|G_{i,:}\|_2^2\Big)^{1/2}=\frac{\|G\|_F}{\sqrt m},
\qquad c=\frac{\rho}{1-\rho}\,\bar g,
\]
with a unitless $\rho\in[0,1]$.  
Since $\mathsf{Spectral}_s$ is invariant
to positive scaling, we can replace $\mathsf{RowClip}_c(G)$ by
\begin{equation}
\tilde G_{i,:}=\frac{G_{i,:}}{\max\{(1-\rho)\|G_{i,:}\|_2,\ \rho\bar g\}}.
\label{eq:sparrow-rho}
\end{equation}
We make this change for numerical stability, since this form is defined on
all of $[0,1]$ and attains both ends exactly:
\begin{itemize}
    \item $\rho=0$: $\tilde G=\mathsf{Row}_1(G)$, which gives \Sparrow{} ($U_3$).
    \item $\rho=\tfrac12$: $c=\bar g$, so only the rows longer than the
    root-mean-square row norm are clipped.
    \item $\rho=1$: $\tilde G=G/\bar g$, so no row is clipped, which gives
    $U_1$ in \eqref{eq:rowclip-polar-approximation}.
\end{itemize}
It also keeps the input of $\mathsf{Spectral}_s$ at a scale that does not
depend on $G$.
Like \Sparrow{}, \Cref{alg:sparrow-c}
applies $\mathsf{Spectral}_s$ at most once and its output lies in
$\mathcal K_{r,s}$.
The parameter $\rho$ need not be fixed during training; one could, for instance,
schedule it inversely proportional to the step-size schedule $\gamma_k$, so that the
update is close to \Sparrow{} at the peak step size and moves towards $U_1$ as the
step size decays.

\begin{algorithm}[H]
\caption{\textsf{Sparrow}$_\rho$ update direction.  For $\rho=0$ it is \Sparrow{} and for $\rho=1$ it is $U_1$.}
\label{alg:sparrow-c}
\begin{algorithmic}[1]
\renewcommand{\algorithmicensure}{\textbf{Return:}}
\Require gradient estimate $G\in\mathbb R^{m\times d}$, radii $s$ and $r=s\min\{1,\sqrt{d/m}\}$, $\rho\in[0,1]$
\Ensure depending on the (hidden) parameter group of $G$:
\settowidth{\sparrowlabelwidth}{SwiGLU gate and value matrices ($m>d$):}
\State \makebox[\sparrowlabelwidth][l]{SwiGLU gate and value matrices ($m>d$):}\quad $(\mathsf{RowClip}_r\circ\mathsf{Spectral}_s)(\tilde G)$ with $\tilde G$ in \eqref{eq:sparrow-rho}
\State \makebox[\sparrowlabelwidth][l]{Attention-gate/SSM-gate matrices:}\quad $\mathsf{Row}_r(G)$
\State \makebox[\sparrowlabelwidth][l]{All other matrices:}\quad $\mathsf{Spectral}_s(G)$
\end{algorithmic}
\end{algorithm}

\subsection{Solvers for the joint LMO}
\label[appendix]{app:lmo-idrs}

If our coarse approximation used in \Sparrow{} are not sufficient, one can run an iterative solver like DRS to solve for the joint constraint.
DRS relies on $\mathsf{SpectralClip}_s$ which is typically only computed approximately.  We
therefore compare Douglas--Rachford splitting (DRS) with a fixed approximation
rank against the inexact variant iDRS~\citep{pethick2025ifeddr}, which adapts the rank to the needed accuracy.  The problem is the LMO over
$\mathcal K_{r,s}$, that is, $(a,H)=(0,G)$ in
\Cref{sec:constrained-optimization}.  We find that iDRS certifies every
instance without tuning the rank and is faster than exact DRS throughout
(\Cref{tab:lmo-idrs}).

\paragraph{Problem instances}
The gradients are $G=DU\operatorname{diag}(j^{-\nu})V^\top\in\mathbb
R^{1024\times256}$ with random orthonormal $U,V$, row scales $D$ spaced
geometrically in $[0.2,5]$, and $\|G\|_F=1$.  We take $\nu\in\{1,1.5\}$,
$s=1$, $s/r\in\{6,8,12\}$ and three seeds, which gives 18 instances, all
with both constraints binding.

\paragraph{Douglas--Rachford splitting}
For $\min_W f(W)+g(W)$ with convex $f$ and $g$, DRS iterates, for any
$\gamma>0$,
\begin{equation}
W:=\operatorname{prox}_{\gamma g}(Z),\qquad
Z\leftarrow Z+\operatorname{prox}_{\gamma f}(2W-Z)-W,
\tag{DRS}
\label{eq:drs}
\end{equation}
and $W$ converges to a minimizer.  With $f$ and $g$ from
\Cref{sec:constrained-optimization} and $\gamma=1$,
$\operatorname{prox}_{\gamma f}(Z)=\mathsf{RowClip}_r(Z+\gamma G)$ is exact,
and $\operatorname{prox}_{\gamma g}=\mathsf{SpectralClip}_s$ is the step we
approximate.

\paragraph{Approximate spectral clip}
We clip only the top $k$ singular values, computed on a subspace
$B\in\mathbb R^{d\times k}$ that is warm-started across iterations, and
rescale the remainder so that the output stays feasible:
\[
\begin{aligned}
&Q=\operatorname{qr}(ZB),\qquad
Q^\top Z=\hat U\operatorname{diag}(\alpha)\hat V^\top,\qquad
P=Q\hat U,\\
&T=Z(I-\hat V\hat V^\top),\qquad
c=\min\{1,s/\|T\|_F\},\\
&w=P\operatorname{diag}(\min\{\alpha,s\})\hat V^\top+cT,\qquad
n=\tfrac1\gamma P\operatorname{diag}((\alpha-s)_+)\hat V^\top.
\end{aligned}
\]
Here $w$ is the approximate clip and $n$ lies in the normal cone of the
spectral ball at $w$ for any $B$.  The error of this resolvent step is
\[
\varepsilon=\tfrac1\gamma(Z-w)-n
=\underbrace{\tfrac1\gamma(I-QQ^\top)Z\hat V\hat V^\top}_{\varepsilon_{\mathrm{cross}}}
+\underbrace{\tfrac1\gamma(1-c)T}_{\varepsilon_{\mathrm{tail}}},
\]
where $\varepsilon_{\mathrm{cross}}$ is reduced by sweeping the subspace,
$B\gets\hat V$, and $\varepsilon_{\mathrm{tail}}$ only by increasing $k$.  We
write $(w^t,n^t,\varepsilon^t)$ for the output at iterate $Z^t$.

\paragraph{Fixed-rank DRS}
DRS with a fixed $k\in\{16,32,64,128\}$ replaces $W$ in \eqref{eq:drs} by
$\hat W^t=Z^t-\gamma n^t$ and sweeps after every step.  Exact DRS uses a full
SVD, so that $\varepsilon^t=0$.  A fixed rank must be chosen large enough in
advance, since otherwise $\varepsilon_{\mathrm{tail}}$ stalls the iteration.

\paragraph{iDRS}
iDRS~\citep[Alg.~II]{pethick2025ifeddr} starts at $k=4$ and, at iterate
$Z^t$, forms
\[
\tilde Z^t=w^t-\gamma n^t,\qquad
Y^t=\tfrac1\gamma\big(\tilde Z^t-\operatorname{prox}_{\gamma f}(\tilde Z^t)\big),\qquad
v^t=n^t+Y^t,\qquad
\bar Z^t=w^t-\gamma Y^t.
\]
If the error is small relative to the residual,
$\|\varepsilon^t\|_F\le0.9\max\{\|v^t\|_F,\|Z^t-\bar Z^t\|_F/\gamma\}$, it
takes the projective step
\[
Z^{t+1}=Z^t-\tfrac{\langle v^t,Z^t-\bar Z^t\rangle}{\|v^t\|_F^2}\,v^t.
\]
Otherwise it recomputes the spectral clip at $Z^t$, after doubling $k$ if
$\|\varepsilon^t_{\mathrm{tail}}\|_F\ge\|\varepsilon^t_{\mathrm{cross}}\|_F$
and after sweeping the subspace otherwise.

\paragraph{Stopping and timing}
All methods stop once the duality gap of \eqref{eq:lmo-dual} at
$(\mathsf{RowClip}_r(w^t),n^t)$ falls below
\[
10^{-6}\min\{s\|G\|_{\mathrm{nuc}},\,r\|G\|_{2,1}\}.
\]
Timings are
single-threaded NumPy on an Apple-silicon laptop.

\paragraph{Results}
A fixed rank that is too small stalls at a truncation bias, and the rank
needed ranges from 16 to 128 across instances (\Cref{tab:lmo-idrs}).  Over
all 18 instances, iDRS takes 65\,s, against 186\,s for exact DRS and 120\,s
for $k=128$, the smallest fixed rank that certifies every instance.  Choosing
the best rank per instance in hindsight would take 54\,s.  The table also
shows that the cheap composed candidates of
\eqref{eq:rowclip-polar-approximation} lose 20--39\% of the optimal value on
these instances.

\begin{table}[ht]
    \centering
    \small
    \caption{Joint LMO with inexact spectral steps ($1024\times256$).
    Median solve time in seconds over three seeds with iDRS's final rank in
    parentheses.  ``Best-of-three gap'' shows the relative value lost by the best
    composed candidate $U_1,U_2,U_3$ in \eqref{eq:rowclip-polar-approximation}.  ``stall ($g$)'' when DRS fails to
    certify on every seed, with largest final relative gap $g$.}
    \label{tab:lmo-idrs}
    \setlength{\tabcolsep}{4pt}
    \resizebox{\linewidth}{!}{%
    \begin{tabular}{@{}ccccccccc@{}}
        \toprule
        & & & \multicolumn{2}{c}{Adaptive / exact} & \multicolumn{4}{c}{Fixed-rank DRS} \\
        \cmidrule(lr){4-5}\cmidrule(l){6-9}
        $\nu$ & $s/r$ & Best-of-three gap & iDRS & Exact DRS & $k=16$ & $k=32$ & $k=64$ & $k=128$ \\
        \midrule
        1   & 6  & 34\% & \textbf{4.33} (128) & 6.33  & stall (7e-2) & stall (3e-2) & stall (9e-3) & 4.03 \\
        1   & 8  & 29\% & \textbf{1.46} (64)  & 3.76  & stall (4e-2) & stall (1e-2) & 1.20 & 2.40 \\
        1   & 12 & 20\% & \textbf{0.37} (16)  & 1.87  & 0.25 & 0.37 & 0.66 & 1.30 \\
        1.5 & 6  & 39\% & \textbf{10.82} (64) & 30.79 & stall (2e-2) & stall (4e-3) & 9.83 & 19.97 \\
        1.5 & 8  & 37\% & \textbf{3.48} (32)  & 13.82 & stall (5e-3) & 2.38 & 4.08 & 8.75 \\
        1.5 & 12 & 30\% & \textbf{1.03} (16)  & 5.19  & 0.69 & 0.94 & 1.56 & 3.31 \\
        \bottomrule
    \end{tabular}}
\end{table}

\subsection{Spectral clipping via matrix products}
\label[appendix]{app:matmul-spectral}

For a compact SVD $X=U\operatorname{diag}(\sigma_j)V^\top$, the LMO is
$\mathsf{Spectral}_s(X)=sUV^\top$, which Newton--Schulz
iterations~\citep{higham2008functions} and Polar
Express~\citep{amsel2025polar} approximate with matrix products alone, as in
Muon~\citep{jordan2024muon}.  \citet{cesista2025spectralclip} first
expressed $\mathsf{SpectralClip}_s$ through such calls, and
\citet{su2025mclip1,su2025mclip2} derived cheaper expressions.  We use the
form from \citet{su2025mclip2} with two independent calls.
For $X\in\mathbb R^{m\times d}$ with $m\ge d$, define
$P:=\tfrac12\bigl(I+s^{-1}\mathsf{Spectral}_s(X^\top X-s^2I)\bigr)$.
Then
\begin{equation}
\mathsf{SpectralClip}_s(X)
=X-\bigl(X-\mathsf{Spectral}_s(X)\bigr)P.
\label{eq:spectral-clip-oracle}
\end{equation}
In each singular direction, an error in $P$ is weighted by
$|\sigma_j-s|$, so errors near the clipping threshold have little effect
on the output.  A slight excess over $s$ can be removed by
rescaling with a certified upper bound on the spectral norm.

\section{Experimental details}
\label[appendix]{app:experimental-details}
\subsection{Robustness-accuracy pareto frontier}
\label[appendix]{app:classification-details}
\label[appendix]{app:block-details}

\Cref{tab:hyper-robustness} summarizes the settings of the robustness
experiments of \Cref{sec:experiments-robustness}.

\begin{table}[!htb]
    \centering
    \small
    \caption{Hyperparameters of the robustness experiments
    (\Cref{sec:experiments-robustness}).  $B=\budgetmult b_0$ is the certificate,
    and $M$ the momentum of the gradient.}
    \label{tab:hyper-robustness}
    \resizebox{\linewidth}{!}{%
    \begin{tabular}{@{}l|cccc@{}}
        \toprule
        Hyperparameter & SGD + Frobenius & Adam + entrywise & Muon + spectral & Ours + joint \\
        \midrule
        Model & \multicolumn{4}{c}{SwiGLU block, $d=64$, $m=256$, unconstrained linear head} \\
        Teachers & \multicolumn{4}{c}{spread ($\kappa^\star=0.43$), concentrated ($\kappa^\star=0.90$)} \\
        Data & \multicolumn{4}{c}{$x\sim\mathcal N(0,I_d)$, 10 classes, 512 fresh samples per step} \\
        Steps $T$ & \multicolumn{4}{c}{1000 (3000 for the long runs)} \\
        Step-size schedule & \multicolumn{4}{c}{$\eta_t=\eta_0\big(0.1+0.45(1+\cos(\pi t/T))\big)$} \\
        Head & \multicolumn{4}{c}{Adam, step size $10^{-2}$, initialized with $\mathcal N(0,1/d)$ entries} \\
        Block initialization & \multicolumn{4}{c}{$0.02\cdot\mathcal N(0,1)$ entries, projected onto the set} \\
        Seeds & \multicolumn{4}{c}{0, 1, 2} \\
        Evaluation & \multicolumn{4}{c}{4096 fixed held-out samples} \\
        \midrule
        Set of $W_g,W_v$ & Frobenius ball & entrywise box & spectral ball & $\mathcal K_{r,s}$ \\
        Radius & $c_{\mathrm F}=\sqrt B$ & $c_\infty=\sqrt{B/(d\sqrt m)}$ & $s=\sqrt B$ & $r=\sqrt{B\kappa^\star}$, $s=\sqrt{B/\kappa^\star}$ \\
        Set of $W_o$ & Frobenius ball, radius 1 & box, radius $1/\sqrt{dm}$ & \multicolumn{2}{c}{unit spectral ball} \\
        Direction & $c_{\mathrm F}M/\|M\|_F$ & $c_\infty\,\mathrm{adam}(\nabla\mathcal L)$ & $\mathsf{Spectral}_s(M)$ & $\mathsf{RowClip}_r(\mathsf{Spectral}_s(M))$ \\
        Momentum & $\mu=0.9$ & $\beta_1/\beta_2=0.9/0.999$ & \multicolumn{2}{c}{$\mu=0.9$} \\
        Step sizes $\eta_0$ & \multicolumn{2}{c}{$\{0.001,0.003,0.01,0.03,0.1,0.3\}$} & \multicolumn{2}{c}{$\{0.001,0.003,0.01,0.03,0.1\}$} \\
        Budgets $\budgetmult$ & \multicolumn{2}{c}{16 or 64 to 4096; from $\tfrac12$ for seed 0} & \multicolumn{2}{c}{subsets of $\{\tfrac12,1,2,4,16,64\}$} \\
        \bottomrule
    \end{tabular}}
\end{table}

\paragraph{Task}
The teacher's head $H^\star\in\mathbb R^{10\times d}$ is a Gaussian matrix divided by its
spectral norm, so its spectral norm is 1, and labels
are the argmax of the teacher's logits.  The loss is cross-entropy, and the
student's head is trained without a constraint by Adam with step size
$10^{-2}$.

\paragraph{Teachers}
Each teacher is generated from a fixed seed as raw matrices
$(W_o,W_g,W_v)$ and then normalized.  All three are divided by their spectral
norms, $\kappa^\star=\max(\|W_g\|_{2\to\infty},\|W_v\|_{2\to\infty})$ is read
off, and $W_g,W_v$ are scaled by $\sqrt{b_0/\kappa^\star}$ with
$b_0=1/\sqrt d$.  Both then have spectral norm
$\sigma^\star=\sqrt{b_0/\kappa^\star}$, and
$\sigma^\star\max(\rho_g^\star,\rho_v^\star)=b_0$, which makes the teacher roughly nonexpansive, and each
student is trained at a budget $B=\budgetmult b_0$.  The raw matrices are:
\begin{itemize}[leftmargin=2em]
    \item \emph{Spread} ($d=64$, $m=256$): $W_g,W_v$ have i.i.d.\
    $\mathcal N(0,1)$ entries, and $W_o\in\mathbb R^{d\times m}$ is a
    uniformly random matrix with orthonormal rows (from the QR factorization
    of a Gaussian matrix), so all its singular values are $1$.
    \item \emph{Concentrated} ($d=64$, $m=256$): the spread teacher with all
    but the first 8 rows of $W_g$ and $W_v$ set to zero.
\end{itemize}

\paragraph{Training}
Each constrained matrix uses the momentum
$M^{t+1}=\mu M^t+(1-\mu)\nabla_W\mathcal L$ with $\mu=0.9$, on 512 fresh
samples per step for $T=1000$ steps, with step size
$\eta_t=\eta_0\big(0.1+0.45(1+\cos(\pi t/T))\big)$.  Besides
$\mathsf{SpectralClip}_s$ and $\mathsf{RowClip}_r$
(\Cref{sec:constrained-optimization}), the projections onto the Frobenius
ball and the entrywise box are
\[
\mathsf{FrobClip}_{c_{\mathrm F}}(X)=\tfrac{X}{\max\{1,\|X\|_F/c_{\mathrm F}\}},
\qquad
\mathsf{BoxClip}_{c_\infty}(X)_{ij}=\max\{-c_\infty,\min\{X_{ij},c_\infty\}\}.
\]
The updates are
\begin{itemize}[leftmargin=2em]
    \item Muon + spectral ball:
    $W\leftarrow\mathsf{SpectralClip}_s\big(W-\eta\,\mathsf{Spectral}_s(M)\big)$;
    \item ours: $W\leftarrow\mathsf{RowClip}_r\big(\mathsf{SpectralClip}_s(W-\eta Z)\big)$
    with $Z=\mathsf{RowClip}_r\big(\mathsf{Spectral}_s(M)\big)$ from \eqref{eq:rowclip-polar-approximation} (\Cref{app:sparrow-block}
    compares it with the direction in \Sparrow{});
    \item SGD + Frobenius ball: $W\leftarrow\mathsf{FrobClip}_{c_{\mathrm F}}\big(W-\eta\,c_{\mathrm F}M/\|M\|_F\big)$;
    \item Adam + entrywise box:
    $W\leftarrow\mathsf{BoxClip}_{c_\infty}\big(W-\eta\,c_\infty\,\mathrm{adam}(\nabla\mathcal L)\big)$.
\end{itemize}
Every step is thus relative to the radius of its set.  $W_o$ lies in a set of
the same type with worst-case spectral norm $1$: the unit spectral ball,
Frobenius radius $1$, or box radius $1/\sqrt{dm}$.  Accuracies and radii are measured on 4096 fixed held-out samples.

\paragraph{Radii of the sets}
Each set's radius makes its worst-case $\rho\sigma$ equal to $B$:
\begin{itemize}[leftmargin=2em]
    \item Frobenius ball: $c_{\mathrm F}=\sqrt B$;
    \item entrywise box: $c_\infty=\sqrt{B/(d\sqrt m)}$, since $\sup\rho\sigma=c_\infty^2d\sqrt m$;
    \item spectral ball: $s=\sqrt B$;
    \item joint set: $(r,s)=(\sqrt{B\kappa},\sqrt{B/\kappa})$.
\end{itemize}

\paragraph{Certified radius}
The certificate of the trained weights is
\[
L=\|H\|_{\mathrm{op}}\tfrac{\sqrt d}{2}\|W_o\|_{\mathrm{op}}
(L_{\mathrm{SiLU}}\rho_v\sigma_g+\rho_g\sigma_v),
\]
from \eqref{eq:swiglu-lipschitz} with $\opN_{\mathrm{RMS}}$ at $\tau=1$, for
which $\sqrt d\,\omega=\sqrt d/2$.  A correctly classified point is certified
at radius $\mathrm{margin}/(\sqrt2L)$~\citep{tsuzuku2018lipschitz}, where the
margin is the gap between the largest and second-largest logit: each logit
difference $f_y-f_j=(e_y-e_j)^\top f$ is $\sqrt2L$-Lipschitz, so a
perturbation of smaller norm cannot change the prediction.  A misclassified
point has radius $0$, and the median is taken over all 4096 held-out
samples, so it is $0$ below $50\%$ accuracy.

\paragraph{Runs}
\Cref{tab:hyper-robustness} lists the step sizes, budgets and seeds.  The
budgets of each method are chosen to cover its accuracy range.  The 3000-step runs
use seed~0, with peak step sizes $\eta_0\in\{0.01,0.03\}$ for Muon and ours.  The split sweep
of \Cref{fig:ts-robust}c uses seed~0, budgets
$\budgetmult\in\{\tfrac12,1,2,4,16\}$ and peak step sizes
$\eta_0\in\{0.01,0.03\}$.  \Cref{tab:ts-robust} averages the per-seed
frontiers; \Cref{fig:ts-robust}a,b pools the runs of all seeds.

\begin{table}[ht]
    \centering
    \small
    \caption{Robustness: maximal clean accuracy and the frontier
    $R_M(a)$ ($\times10^{-3}$) at the accuracy used in
    \Cref{sec:experiments-robustness}.  Mean $\pm$ standard deviation
    over three seeds (the 3000-step runs use one seed).  ``--'': accuracy not
    reached.}
    \label{tab:ts-robust}
    \setlength{\tabcolsep}{4pt}
    \resizebox{\linewidth}{!}{%
    \begin{tabular}{@{}lcccccc@{}}
        \toprule
        & \multicolumn{4}{c}{Spread teacher ($\kappa^\star=0.43$)} & \multicolumn{2}{c}{Concentrated ($\kappa^\star=0.90$)} \\
        \cmidrule(lr){2-5}\cmidrule(l){6-7}
        & \multicolumn{2}{c}{1000 steps} & \multicolumn{2}{c}{3000 steps} & \multicolumn{2}{c}{1000 steps} \\
        \cmidrule(lr){2-3}\cmidrule(lr){4-5}\cmidrule(l){6-7}
        Method & max.\ accuracy & $R(0.70)$ & max.\ accuracy & $R(0.70)$ & max.\ accuracy & $R(0.85)$ \\
        \midrule
        SGD + Frobenius & $0.688 \pm 0.006$ & -- & $0.769$ & $2.76$ & $0.955 \pm 0.005$ & $3.01 \pm 0.10$ \\
        Adam + entrywise & $0.711 \pm 0.007$ & $2.52 \pm 0.09$ & $0.780$ & $2.98$ & $0.954 \pm 0.003$ & $3.26 \pm 0.04$ \\
        Muon + spectral & $0.771 \pm 0.003$ & $3.97 \pm 0.23$ & $0.835$ & $4.10$ & $0.962 \pm 0.003$ & $5.34 \pm 0.22$ \\
        Ours + row/spectral & $\mathbf{0.793 \pm 0.005}$ & $\mathbf{4.75 \pm 0.18}$ & $\mathbf{0.859}$ & $\mathbf{5.04}$ & $0.964 \pm 0.002$ & $5.25 \pm 0.08$ \\
        \bottomrule
    \end{tabular}}
\end{table}

\paragraph{Why the spectral constraint suffers}
Since $W_g$ is tall, spreading spectral weight evenly over all $m$ units
leaves each row a norm of only $s\sqrt{d/m}$ (\Cref{rem:swiglu-budget}), and
since $\|a\odot b\|_2\le\|a\|_\infty\|b\|_2$ for the hidden activation
$a\odot b$ with $a=\operatorname{SiLU}(W_gu)$ and $b=W_vu$, the
activations are shrunk by the same factor.  
In order to concentrate the activations, the spectrally constrained weight matrix would therefore need to concentrate the mass on a few singular values.  On the spread teacher at
$\budgetmult=\tfrac12$, the spectral ball indeed drives the row-to-spectral
ratio of $W_g$ to $0.97$.  The
spectral ball therefore cannot use added width, which we test next.

\begin{figure}[ht]
    \centering
    \includegraphics[width=\linewidth]{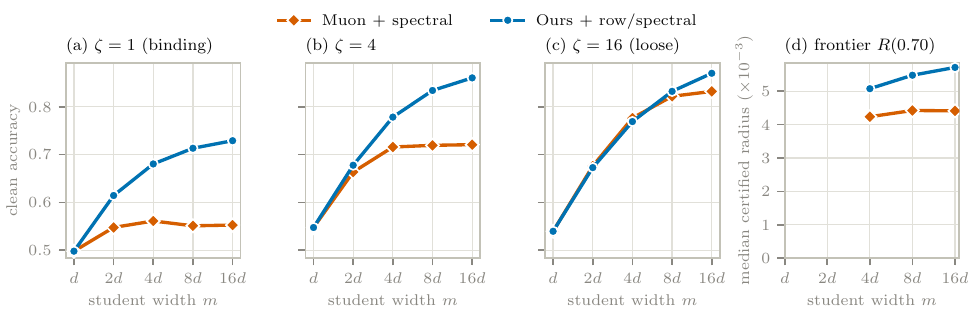}
    \caption{Width sweep at a fixed certificate (teacher $m^\star=16d$).  (a--c)~We observe that the spectral constrained method cannot make use of width when the constraint radius is aggressive.  (d)~The joint constraint certifies a larger radius across widths (at $m\le2d$ no run reaches 70\%).}
    \label{fig:ts-width}
\end{figure}

\paragraph{Varying width}
To test the failure mode of a spectral constraint, we consider Gaussian teacher with a large $m^\star=16d$, normalized as above
($\kappa^\star=0.25$).  Students have width $m\in\{1,2,4,8,16\}\cdot d$ and use
the spectral ball or the joint set with $\kappa=\sqrt{d/m}$, so the two
coincide at $m=d$.  Each student
runs budgets $\budgetmult\in\{1,4,16\}$, peak step sizes
$\eta_0\in\{0.01,0.03,0.1\}$ and seeds
$0,1,2$.  \Cref{fig:ts-width} shows, at each $\budgetmult$, the best clean
accuracy over $\eta_0$ per seed, and the frontier $R(0.70)$ over all
$\budgetmult$ and $\eta_0$, both averaged over seeds.
At $\budgetmult=1$, the accuracy of the spectral ball stays flat as the
student width grows from $4d$ to $16d$ ($0.561$ to $0.553$), whereas the joint
set with $\kappa=\sqrt{d/m}$ improves from $0.680$ to $0.729$.  At
$\budgetmult=16$, where the budget no longer binds, the plateau disappears.

\subsubsection{\Sparrow{} as the direction of the projected step}
\label[appendix]{app:sparrow-block}

\begin{figure}[t]
    \centering
    \includegraphics[width=\linewidth]{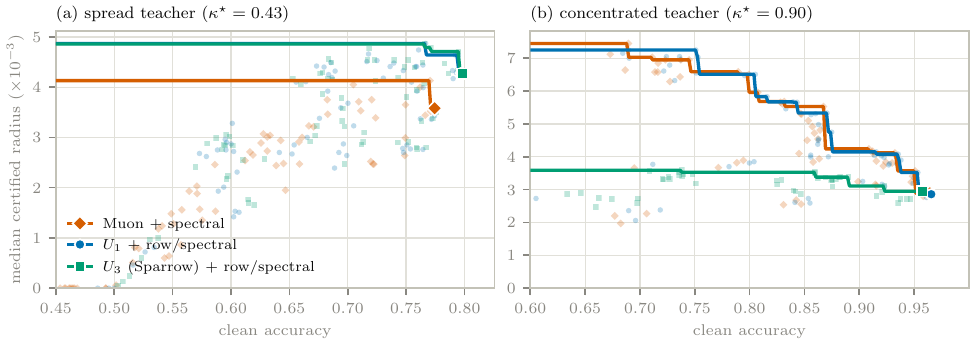}
    \caption{Robustness frontiers $R_M(a)$ of the joint set with the
    direction $U_1$, \Sparrow{}'s $U_3$, and Muon. The two directions coincide on the spread teacher, while $U_3$
    falls behind on the concentrated one.}
    \label{fig:ts-sparrow}
\end{figure}

The robustness experiments use the direction
$U_1=(\mathsf{RowClip}_r\circ\mathsf{Spectral}_s)(M)$ of
\eqref{eq:rowclip-polar-approximation}.  
In contrast, \Sparrow{} uses
$U_3=(\mathsf{RowClip}_r\circ\mathsf{Spectral}_s\circ\mathsf{Row}_r)(M)$.  \Cref{fig:ts-sparrow} compares the
two methods using the same budgets, step sizes and seeds as in
\Cref{sec:experiments-robustness}.
For consistency, all
directions use knowledge of the teacher's split, $r=\sqrt{B\kappa^\star}$ and
$s=\sqrt{B/\kappa^\star}$, instead of the default
$r=s\sqrt{d/m}$ in \Sparrow{}.

On the spread teacher the two directions performs practically identically, beating Muon.
On the concentrated teacher, however, $U_3$ performs significantly worse than $U_1$ and Muon.
One could potentially use this as a practical diagnostic.  
If \Sparrow{} performs worse than Muon in practice, the target might be concentrated on few hidden units.
Since a model of adequate size should use its hidden units densely, this in
turn suggests that the model is wider than necessary.

\subsection{\Sparrow{} as a drop-in replacement for Muon}
\label[appendix]{app:speed-details}

\begin{table}[!htb]
    \centering
    \small
    \caption{Hyperparameters of the language-model experiments
    (\Cref{sec:experiments-speed}).  The methods differ
    only in the direction of the listed matrices. ``\Sparrow{} (gates only)''
    in \Cref{fig:gate} uses Muon on $W_g,W_v$ and \Sparrow{} on $W_a$.  Step sizes
    list the values tried for each method (best bolded).}
    \label{tab:hyper-nanogpt}
    \resizebox{\linewidth}{!}{%
    \begin{tabular}{@{}l|ccc@{}}
        \toprule
        Hyperparameter & Muon & Aurora & \Sparrow{} \\
        \midrule
        Dataset & \multicolumn{3}{c}{Tiny Shakespeare, characters, first $90\%$ for training} \\
        Layers / heads / width $d$ & \multicolumn{3}{c}{4 / 4 / 128} \\
        Head width & \multicolumn{3}{c}{32 (tall gate: 128)} \\
        MLP / normalization & \multicolumn{3}{c}{SwiGLU with $m=512$ / RMSNorm} \\
        Context / batch size & \multicolumn{3}{c}{64 / 12} \\
        Steps & \multicolumn{3}{c}{1307 (one epoch)} \\
        Step-size schedule & \multicolumn{3}{c}{100 warmup steps, then cosine decay to $10\%$} \\
        Gradient clipping & \multicolumn{3}{c}{global norm 1} \\
        Embeddings and head & \multicolumn{3}{c}{AdamW, step size $10^{-3}$, $\beta_2=0.99$, weight decay $0.1$} \\
        Initialization & \multicolumn{3}{c}{std $0.02$, residual projections $0.02/\sqrt{8}$} \\
        Hidden matrices & \multicolumn{3}{c}{$W\leftarrow W-\eta\lmo(M)$, no projection, no weight decay} \\
        Momentum & \multicolumn{3}{c}{0.9, without Nesterov} \\
        $\mathsf{Spectral}_s$ & \multicolumn{3}{c}{exact, via SVD (Polar Express in \Cref{fig:polar-approx})} \\
        Seeds & \multicolumn{3}{c}{10} \\
        Evaluation & \multicolumn{3}{c}{every 100 steps, 20 batches of the last $10\%$} \\
        \midrule
        Step size $\eta$ (\Cref{fig:speed}) & $\{0.005,\mathbf{0.01},0.02\}$ & $\{0.01,\mathbf{0.02},0.04\}$ & $\{0.01,\mathbf{0.02}\}$ \\
        Step size $\eta$ (\Cref{fig:gate}) & $0.01$ & $0.01$ & $0.01$ \\
        $\lmo$ of $W_g,W_v$ & $\mathsf{Spectral}_1$ & \multicolumn{2}{c}{Aurora / \Sparrow{}, $s=1$, $r=s\sqrt{d/m}=\tfrac12$} \\
        $\lmo$ of gate $W_a$ & $\mathsf{Spectral}_1$ & Aurora if tall & $\mathsf{Row}_r$, $r=s\min\{1,\sqrt{d/(Hd_v)}\}\in\{1,\tfrac12\}$ \\
        $\lmo$ of other matrices & $\mathsf{Spectral}_1$ & Aurora if tall, else $\mathsf{Spectral}_1$ & $\mathsf{Spectral}_1$ \\
        \midrule
        Weight normalization (hatched) & MuonH & -- & SparrowH; without normalized direction \\
        \quad initialization / radius & \multicolumn{3}{c}{std $1/\sqrt{n}$ / $R=\|W_0\|_F$} \\
        \quad step size $\eta$ & $\{0.005,\mathbf{0.01},0.02,0.04,0.08\}$ & -- & $0.01$ \\
        \bottomrule
    \end{tabular}}
\end{table}

\Cref{tab:hyper-nanogpt} lists the model and the optimizers.  The training
split, the first $90\%$ of Tiny Shakespeare, is cut into non-overlapping
windows of 64 characters, which are shuffled and each used once, so a run of
1307 steps sees every training character once.

\paragraph{Aurora}
Aurora is Algorithm~3 of~\citet{dewulf2026aurora} with their default of two iterations and damping
$\beta=\tfrac12$: from $X_0=M/\|M\|_F$ and $D_0=I$, it computes, for $k=0,1$,
\begin{equation*}
D_{k+1}=D_k^{\beta}\operatorname{diag}\big(\|(X_k)_{i,:}\|_2\big)^{1-\beta},\qquad
X_{k+1}=\mathsf{Spectral}_1\big(D_{k+1}^{-1}X_k\big),
\end{equation*}
and returns $sX_2$.

\paragraph{Speed and cost}
Each method is compared
with Muon on the same seed which fixes both the initialization and the data order.  
We measure speed by the first step at which a run reaches Muon's final validation loss on the same seed. Muon's own validation loss is not monotone, in fact on four seeds it reaches its final value earlier, which makes its mean 1302 rather than 1307.
Computing a single direction for a $512\times128$ matrix takes $2.6$\,ms
for Muon and \Sparrow{} and $5.2$\,ms for Aurora (mean of 200 compiled calls).

\begin{figure}[t]
    \centering
    \begin{minipage}[t]{0.485\linewidth}
        \centering
        \includegraphics[width=\linewidth]{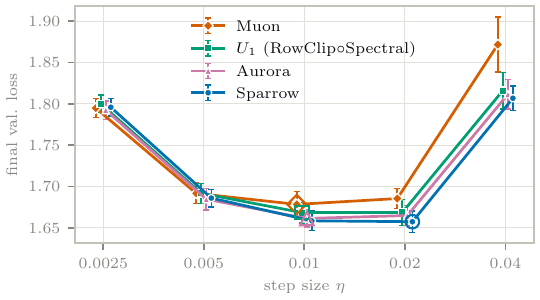}
        \caption{Step-size sweep on Tiny Shakespeare. 
        Final validation loss of
        Muon, $U_1=(\mathsf{RowClip}_r\circ\mathsf{Spectral}_s)(M)$, Aurora and \Sparrow{},
        which differ only on $W_g$ and $W_v$.  Mean $\pm$ one standard error over
        three seeds with the best step size of each method being circled.
        \Sparrow{} seems to allow for slightly larger stepsize.}
        \label{fig:speed-lr}
    \end{minipage}\hfill
    \begin{minipage}[t]{0.485\linewidth}
        \centering
        \includegraphics[width=\linewidth]{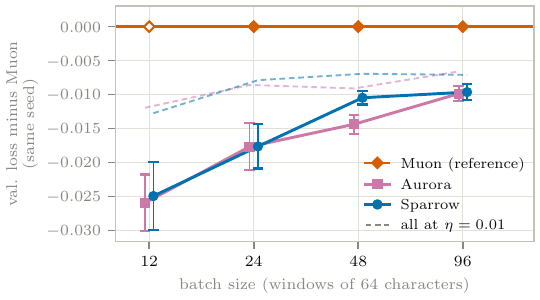}
        \caption{Batch-size sweep on the Complete Works of Shakespeare, 750 steps
        without repeated data. Each method uses its own best step
        size per batch size (filled markers: $\eta=0.02$, hollow:
        $\eta=0.01$). 
        Mean $\pm$ one standard error over three seeds. \Sparrow{} is beneficial across batch sizes and particularly at small batch sizes, on par with Aurora at half the cost of its LMO.}
        \label{fig:batch-size}
    \end{minipage}
\end{figure}

\paragraph{Step-size sweep}
\Cref{fig:speed-lr} repeats the comparison over
$\eta\in\{0.0025,0.005,0.01,0.02,0.04\}$ for Muon, $U_1$ of
\eqref{eq:rowclip-polar-approximation}, Aurora and \Sparrow{} on $W_g,W_v$, with seeds $0,1,2$.

\begin{figure}[t]
    \centering
    \includegraphics[width=\linewidth]{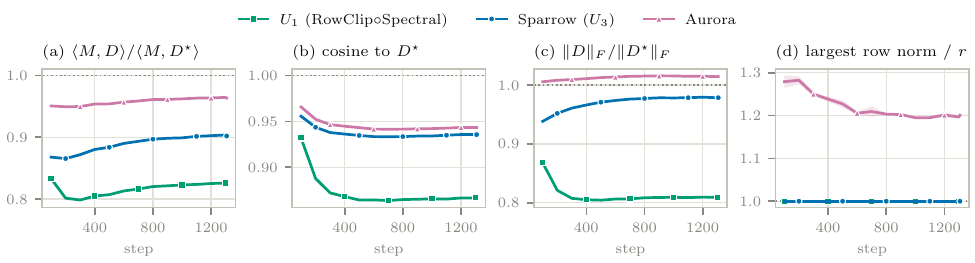}
    \caption{Approximate oracles against the joint LMO $D^\star$ of $\mathcal K_{r,s}$ on the momentum of the
    SwiGLU gate and value matrices along Muon's training runs (mean over the eight matrices, mean $\pm$ one
    standard error over three seeds).  $U_1$ and \Sparrow{} coincide in (d).}
    \label{fig:lmo-exact}
\end{figure}

\paragraph{Comparison with the exact LMO}
Along Muon's runs ($\eta=0.01$, seeds $0,1,2$), we map the momentum $M$ of
each $W_g,W_v$ by $U_1$, \Sparrow{} ($U_3$) and Aurora, and compare
each direction $D$ with the joint LMO $D^\star$ of $\mathcal K_{r,s}$, computed
by \ref{eq:drs} to a relative duality gap below
$1.1\cdot10^{-3}$ (\Cref{fig:lmo-exact}).  \Sparrow{} attains $87$--$90\%$ of
the optimal value $\langle M,D^\star\rangle$, against $80$--$83\%$ for $U_1$.
Aurora attains $95$--$96\%$, but leaves the set, with a largest row norm
$20$--$28\%$ above $r$.  Importantly, the directions of \Sparrow{} and Aurora are close
(cosine to $D^\star$ of $0.93$ and $0.94$ after the first few hundred steps),
so most of \Sparrow{} approximation error is due to length ($2$--$6\%$ below that of
$D^\star$), which a larger step size compensates for (\Cref{fig:speed-lr}).
\looseness=-1

\paragraph{Batch size}
To vary the batch size without repeating data, \Cref{fig:batch-size} uses the
Complete Works of Shakespeare ($75{,}365$ training windows) and trains for
750 steps at batch $12,24,48$ and $96$, scoring every run on the same
$20\times96$ validation windows.  Each method uses its own best step size per
batch size from $\eta\in\{0.01,0.02\}$ on seeds $0,1,2$, with $\eta=0.04$
added at batch 96, where it loses for all three methods, and $\eta=0.005$
for Muon at batch 12.  
The best step size is $0.02$ everywhere except for
Muon at batch 12, where it is $0.01$.
The improvement of \Sparrow{} over Muon interesting seems to be larger at small batch sizes.  

\begin{figure}[H]
    \centering
    \includegraphics[width=0.55\linewidth]{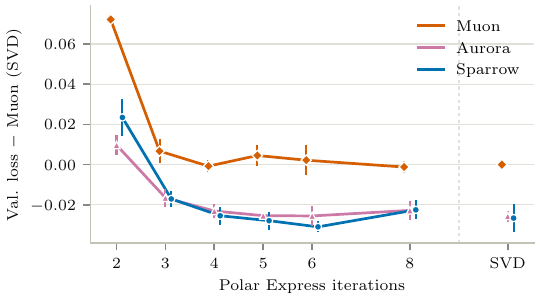}
    \caption{Inexact polar factor on Tiny Shakespeare at $\eta=0.02$.  Final
    validation loss minus that of Muon with the exact SVD on the same seed, when
    every polar factor is computed by $k$ iterations of Polar
    Express~\citep{amsel2025polar} and the rightmost points use the SVD.  Mean $\pm$
    one standard error of the paired difference over three seeds.}
    \label{fig:polar-approx}
\end{figure}

\paragraph{Inexact polar factor}
All other runs compute $\mathsf{Spectral}_s$ exactly by an SVD.  In
\Cref{fig:polar-approx}, every polar factor is instead computed by $k$ iterations of Polar
Express~\citep{amsel2025polar}, at $\eta=0.02$ for all three methods with seeds $0,1,2$.  We observe that \Sparrow{} is
robust to this approximation of $\mathsf{Spectral}_s$.

\paragraph{Gated attention}
The gate multiplies the attention output before $W_O$ channelwise by
$\sigma(\opN(z)W_a^\top)$, with $W_a\in\mathbb R^{Hd_v\times d}$, no bias,
$c=1$ and $\opN$ the RMSNorm input of the attention block.  $W_a$ is
initialized from its own random key, so the other weights match the ungated
model. Note that the head width 128 of the tall gate also makes $W_Q,W_K,W_V$ tall.
\Cref{tab:hyper-nanogpt} gives the directions and radii of each method.

\paragraph{Weight normalization}
For the hatched bars of \Cref{fig:gate}, every hidden matrix $W\in\mathbb
R^{m\times n}$ is initialized with entries of standard deviation $1/\sqrt n$
and kept at $R=\|W_0\|_F$ by $\mathsf N$ of
\eqref{eq:frobenius-normalization}~\citep{wen2026hyperball}.  With $D$ the
direction of Muon or \Sparrow{}, MuonH and SparrowH take
$W\leftarrow\mathsf N(W-\eta\,\mathsf N(D))$, and \Sparrow{} without
normalized direction takes
$W\leftarrow\mathsf N\big(W-\eta\tfrac{R}{s\sqrt{\min\{m,n\}}}D\big)$. 
Since the initialization
differs from the other runs, the effect of \Sparrow{} under weight
normalization should be compared against MuonH.

\subsection{Looped models}
\label[appendix]{app:looped-shakespeare}

The looped model of \Cref{sec:experiments-looped} is the tall-gate model of \Cref{tab:hyper-nanogpt}.  The first layer maps the embedding to the anchor $u$, the second is iterated $N$ times,
and the third reads out $z^N$:
\begin{equation*}
\begin{aligned}
\text{prelude:}\quad & u=(\opI+\opM_1)\circ(\opI+\opA_1)(E_x+P),\\
\text{loop:}\quad & z^{k+1}=\tfrac12u+\tfrac12\opT(z^k),\qquad z^0=u,\quad k<N,\\
\text{core:}\quad & \opT=(\opI+\opM_2)\circ(\opI+\opA_2),\\
\text{coda:}\quad & \text{logits}=\opN\big((\opI+\opM_3)\circ(\opI+\opA_3)(z^N)\big)W_{\mathrm{head}}^\top,
\end{aligned}
\end{equation*}
where $E_x=(E_{x_1},\dots,E_{x_T})$ are the rows of the learned token embedding $E$ for the input characters and
$P$ the learned position embeddings, $\opA_\ell$ is the gated attention and $\opM_\ell$ the
SwiGLU block \eqref{eq:swiglu-map} of layer $\ell$, both with the RMSNorm $\opN$ on their input.  Truncated
backpropagation detaches $z^{N-H}$ (the gradient still flows to the first layer through the anchor $u$).  \Sparrow{} is applied to all matrices, as in \Sparrow{} (all) of
\Cref{fig:gate}.  Each method uses its best step size of $\{0.01,0.02\}$ per configuration across three seeds: $0.01$
for Muon (except $0.02$ at $H=8$) and $0.02$ for \Sparrow{} (except $0.01$ at $N=1$).

\end{document}